\documentclass[final,1p,times]{elsarticle}

\usepackage{amssymb}
\usepackage{amssymb}
\usepackage{float}
\usepackage {amsmath}
\usepackage{algorithm}  
\usepackage{algpseudocode}  
\usepackage{amsmath}  
\usepackage{booktabs} 
\usepackage{caption} 
\usepackage{setspace}
\usepackage{subcaption}
\usepackage{hyperref}
\usepackage{threeparttable}
\usepackage{multirow}
\usepackage{doi}
\usepackage{xcolor} 
\usepackage{caption}

\hypersetup{
    colorlinks=true, 
    linkcolor=blue,  
    urlcolor=blue,   
    citecolor=blue   
}

\makeatletter
\renewcommand{\fnum@figure}{\textbf{Fig. \thefigure.}\@gobble}
\makeatother

\makeatletter
\renewcommand{\fnum@table}{\textbf{Table \thetable }\@gobble}
\makeatother

\journal{Elsevier}

\begin{document}

\begin{frontmatter}



\title{Dual Adversarial and Contrastive Network for Single-Source Domain Generalization in Fault Diagnosis}


\author[a]{Guangqiang Li} 
\author[b]{M. Amine Atoui} 
\author[a]{Xiangshun Li\corref{cor1}} 
\ead{lixiangshun@whut.edu.cn}
\cortext[cor1]{Corresponding authors at: School of Automation, Wuhan University of Technology, Wuhan 430070, PR China.}

\affiliation[a]{organization={School of Automation},
            addressline={Wuhan University of Technology}, 
            city={Wuhan},
            postcode={430070}, 
            country={PR China}}
\affiliation[b]{organization={The School of Information Technology},
            addressline={Halmstad University}, 
            city={Halmstad},
            country={Sweden}}

\begin{abstract}
Domain generalization achieves fault diagnosis on unseen modes. {In process industrial systems, fault samples are limited, and it is quite common that the available fault data are from a single mode.} Extracting domain-invariant features from single-mode data for unseen mode fault diagnosis poses challenges. Existing methods utilize a generator module to simulate samples of unseen modes. However, multi-mode samples contain complex spatiotemporal information, which brings significant difficulties to accurate sample generation.
To solve this problem, this paper proposed a dual adversarial and contrastive network (DACN) for single-source domain generalization in fault diagnosis. The main idea of DACN is to generate diverse sample features and extract domain-invariant feature representations. 
An adversarial pseudo-sample feature generation strategy is developed to create fake unseen mode sample features with sufficient semantic information and diversity, leveraging adversarial learning between the feature transformer and domain-invariant feature extractor. An enhanced domain-invariant feature extraction strategy is designed to capture common feature representations across multi-modes, utilizing contrastive learning and adversarial learning between the domain-invariant feature extractor and the discriminator.
Experiments on the Tennessee Eastman process and continuous stirred-tank reactor demonstrate that DACN achieves high classification accuracy on unseen modes while maintaining a small model size.

\end{abstract}


\begin{highlights}
\item Dual adversarial and contrastive network (DACN) is proposed for single-source domain generalization in multi-mode fault diagnosis.
\item {An adversarial pseudo-sample feature generation strategy is developed to generate fake unseen mode sample features with sufficient semantic information and diversity.}
\item {An enhanced domain-invariant feature extraction strategy is designed to capture common feature representations of system health conditions across different modes.}
\item The experiments demonstrate DACN's robust generalization and compact model size.

\end{highlights}


\end{frontmatter}



\section{Introduction}
Modern process systems are highly automated and intelligent, resulting in substantial increases in scale and complexity \cite{RN192229}. Within the plant system, the coupling of equipment is strong. Fault of a single equipment can cause a series of equipment faults through internal propagation, or even lead to a loss of control of the system. Faulty events have significant economic, safety and environmental impact \cite{RN198691}. The type, location, magnitude, and time of the fault can be determined by fault diagnosis model \cite{RN237009}. Therefore, fault diagnosis plays an important role in ensuring the safe and reliable operation of process systems \cite{RN191405}. 

Data-driven fault diagnosis methods have gained widespread attention due to their high accuracy and low requirement for specialized knowledge \cite{RN225608,RN214605}. Data-based methods can be divided into multivariate statistical analysis based and machine learning based methods. Multivariate statistical analysis based methods have been successfully applied to industrial systems, such as principal component analysis (PCA), independent component analysis (ICA), partial least squares (PLS) and Fisher's discriminant analysis \cite{RN198951,RN233663,RN233068,RN233872}. This kind of methods usually have specific assumptions, which makes them not applicable to all situations. Among methods based on machine learning, k-nearest neighbors (KNN), support vector machine (SVM), random forest (RF) and Bayesian network (BN) have been successfully applied \cite{RN237010,RN230344,RN190802,RN214728,RN191413}. Lou et al. proposed statistical subspaces to deal with unseen faults \cite{RN237011}. Deng et al. proposed a fault detection method based on space-time compressed matrix and naive Bayes, which can significantly reduce learning complexity while ensuring classification performance \cite{RN215033}. These methods are limited in processing high-dimensional data, and the performance depends on the quality of feature engineering.

Compared with traditional machine learning methods, deep learning based methods can automatically extract features and achieve high accuracy \cite{RN237012}. Convolutional neural network (CNN), deep belief network (DBN), stacked autoencoder (SAE), long short-term memory network (LSTM), and vision transformer (ViT) have been successfully applied to solve the fault diagnosis problems in process industry systems \cite{RN194629,RN195057,RN237013,RN237014,RN237015, lou2023unknown}. Hashim et al. combines a multi-Gaussian assumption and an attribute fusion network to enhance zero-shot fault diagnosis \cite{RN237035}. Huang et al. considered time delay of fault occurrence and proposed the CNN-LSTM model \cite{RN237022}. Zhang et al. developed a maximum smooth function (MSF) to replace the classical activation function and proposed a gated recurrent unit-enhanced deep convolutional neural network model \cite{RN187493}. Wei et al. introduced target-attention in the decoder of the transformer \cite{RN189089}. This method enhanced the ability to capture long-term dependencies and detect faults early. Ren et al. designed generators in the amplitude-frequency and phase-frequency domains, utilizing the Pearson correlation coefficient and a time domain discriminator to generate samples with reliability, authenticity, and diversity \cite{RN255604}. Gao et al. proposed a semi-pseudo-labeling diagnosis ResNet (SPL-DResNet) system that enhances weak-fault diagnosis by evaluating sample reliability to mitigate label noise in pseudo-labels, thereby improving the extraction of fault-related features \cite{RN255606}.

Deep learning based methods \cite{lou2023recent} assume that the training set and the test set follow the same data distribution. The operating points of industrial systems vary with the environment, loads and production requirements, resulting in differences in the data distribution between the training and test sets \cite{RN228668}. Transfer learning aims to extract the knowledge from one or more source tasks and applies the knowledge to a target task, which can solve the problem of changes in data distribution \cite{RN237023}. For labeled and unlabeled fault data of target mode, Wu et al. proposed fine-tuning and joint adaptive network methods to build a fault diagnosis model \cite{RN216669}. Wang et al. used linear discriminant analysis (LDA) to assign weights to each feature variable and designed a weighted maximum mean discrepancy (MMD) for domain adaptation \cite{RN192444}. Gao et al. used class prior distribution adaptation to generate class-specific auxiliary weights. This method can exploit the prior probability on the source and target domains \cite{RN189748}. Chai et al. suggested to learn features through multi-domain discriminators competition. This method achieved global alignment of two domains and fine-grained alignment of each fault class \cite{RN192669}. Gao et al. proposed a time-frequency supervised contrastive learning framework (TF-SupCon) for bearing fault diagnosis under variable speed conditions, leveraging supervised contrastive learning and KNN-based pseudo-labeling to enhance the extraction of speed-invariant features and improve cross-domain diagnostic performance\cite{RN255605}. Transfer learning based methods require both labeled source domain data and unlabeled target domain data. However, the target domain data is usually unavailable in real systems. 

Domain generalization intends to learn a model from one or several domains that can be generalized to unseen domains \cite{RN237017}. Domain generalization based methods have been successfully applied in fault diagnosis. Xiao et al. proposed to align the label distribution by weighting the source classification loss, and to learn domain-invariant feature representations by progressive adversarial training \cite{RN188446}. Huang et al. constructed causal layers to discover causal mechanism in features and developed a domain classifier cooperating with gradient reversal layers to capture the domain invariance \cite{RN213793}. Xiao et al. proposed a method to learn a classifier and invariant feature representation simultaneously based on the weighted variables \cite{RN190479}. Domain generalization based methods learn fault knowledge in multi-mode data that can be generalized to unseen modes. {However, fault samples from multi-mode are difficult to collect , and they are usually obtained from a single mode.}

In single-source domain generalization for multi-mode fault diagnosis, a common approach is to first generate unseen mode samples from source domain samples, then construct a fault diagnosis model using both source and generated samples. Zhao et al. proposed an adversarial mutual information-guided single domain generalization network (AMINet) \cite{RN187888}. This method designs an iterative min–max game of mutual information between the domain generation module and task diagnosis module to learn generalized features. Wang et al. designed an adversarial contrastive learning strategy, which can promote the learning of class-wise domain-invariant representations while maintaining the diversity of the generated samples \cite{RN186301}. Pu et al. presented an incremental domain augmentation strategy. This method can generate several augmented domains with different distributions but the same semantic information \cite{RN237026}. Guo et al. used Mixup to create augmented domains with significant distributional differences from the source domain \cite{RN237028}. However, the data collected from industrial systems contain complex spatial-temporal information, accurately generating samples of unknown modes is challenging. Additionally, most methods use classifiers as arbiters to determine the system health condition class of the generated samples. This can lead to erroneous samples being generated by the generator before the classifier has fully learned the true classification criteria.

To address the above problems, this paper proposes a novel network architecture, the dual adversarial and contrastive network (DACN). {The DACN model is designed with a feature transformer module and a domain invariant feature extractor module for generating diverse sample features and extracting common features across multiple modes, respectively. The DACN is pretrained to learn the feature representations of single seen mode samples related to the system health condition. Feature transformer is employed to generate pseudo-sample features based on the feature representations of single seen mode samples. To enhance the diversity of pseudo-sample features	and the cross-mode common feature representations, an adversarial pseudo-sample feature generation strategy and an enhanced domain-invariant feature extraction strategy are designed. The adversarial pseudo-sample feature generation strategy is constructed to train the feature transformer module, increasing the distributional difference between pseudo-sample features and single seen mode sample features. The enhanced domain-invariant feature extraction strategy is developed to learn common discriminative representations from pseudo-sample features and single seen mode sample features. After the generation of diverse pseudo-sample features and the learning of domain-invariant feature representations, the generalization ability of DACN against distribution shifts in unseen modes is enhanced.}

The main contributions of this paper are as follows: (1) DACN is proposed to solve the problem of single-domain generalization in fault diagnosis. The model can be trained from a single mode of monitoring data and generalized to multiple unseen modes. (2) An adversarial pseudo-sample feature generation strategy is developed to generate fake unseen mode sample features with sufficient semantic information and diversity, avoiding complex sample generation processes. (3) An enhanced domain-invariant feature extraction strategy is designed to capture common representations of system health condition across multiple modes, enabling robust fault diagnosis across different modes. (4) The effectiveness of DACN is validated through extensive experiments on the Tennessee Eastman process and the Continuous Stirred-Tank Reactor (CSTR). The results indicate that DACN not only achieves robust generalization but also maintains a compact model size.

{The rest of this paper is organized as follows. \hyperref[Sec2]{Section \ref{Sec2}} introduces the problem description of single-source domain generalization in fault diagnosis and the motivation of this study. \hyperref[Sec3]{Section \ref{Sec3}} shows the detailed structure and training process of the DACN model. \hyperref[Sec4]{Section \ref{Sec4}} describes the experimental setup. \hyperref[Sec5]{Section \ref{Sec5}} presents the experimental results. Finally, conclusions are drawn in \hyperref[Sec6]{Section \ref{Sec6}}.}

\section{Preliminaries}
\label{Sec2}
\subsection{Problem formulation}

{A multi-mode process refers to a process that exhibits multiple stable operating states, which are influenced by external conditions, production schedules, or inherent characteristics of the process. Each stable operating state is referred to as a single mode. In the single-source domain generalization for fault diagnosis scenario, the monitoring data from a single seen mode constitute the source domain; while data from other unseen modes constitute the target domain. The data in the source domain is used to train the fault diagnosis model, while the data in the target domain is not accessible during the training phase and is only used to evaluate the generalization performance of the model.
Let $D_{\mathrm{S}}=\{x_{i}^{\mathrm{S}},y_{i}^{\mathrm{S}}\}$ represent the source domain, where $i=1,...,N_{\mathrm{S}}$, $N_{\mathrm{S}}$ is the number of samples in the source domain, $x_{i}^{\mathrm{S}} \in \mathbb{R}^{v}$ denotes the $i$-th sample and $y_{i}^{\mathrm{S}} \in \{1, 2, ..., L\}$ denotes the system health condition label corresponding to the $i$th sample. Here, $v$ is the number of measured variables, and $L$ is the number of system health condition categories, which includes one normal state and $L-1$ fault categories. Similarly, let $D_{\mathrm{T}}=\{x_{j}^{\mathrm{T}}\}$ represent the target domain, where $j=1,...,N_{\mathrm{T}}$, $N_{\mathrm{T}}$ is the number of samples in the target domain, $x_{j}^{\mathrm{T}} \in \mathbb{R}^{v}$ denotes the $j$-th sample in the target domain. 
It is assumed that both the source domain $D_{\mathrm{S}}$ and the target domain $D_{\mathrm{T}}$ share the same feature space and label space, but their data distributions are different, i.e., $P_{\mathrm{T}}\left(x^{\mathrm{T}}\right)\neq P_{\mathrm{S}}\left(x^{\mathrm{S}}\right)$.
The goal of single-source domain generalization in fault diagnosis is to train a model on the source domain, enabling it to effectively generalize to the target domain and accurately inference the health condition of the system in the target domain.}

\subsection{Motivation}
\label{Sec2-2}
The objective of this study is to develop a model that demonstrates robust generalization performance on various unseen modes, utilizing data from a single mode. In the field of image style transfer, images of a specific style contain content and style information \cite{RN237024}. Similarly, the monitoring data for a specific mode can be viewed as a combination of system health condition features and mode features. When the system operates stably in one mode, it can be regarded as a wide-sense stationary process. A wide-sense stationary process can be described by its mean and autocorrelation function. Assume the mode features in the monitoring data are described by the mean of each variable under normal condition. Standardizing each variable by subtracting the mean and dividing by the standard deviation removes mode features and scales the measured variables to the same level. The standardized data primarily reflects the system's health condition features, corresponding to its health condition class.

\begin{figure}[!ht]
    \begin{subfigure}[b]{1\linewidth}
        \centering
		\includegraphics[width=1.\textwidth,height=0.3632\textwidth]{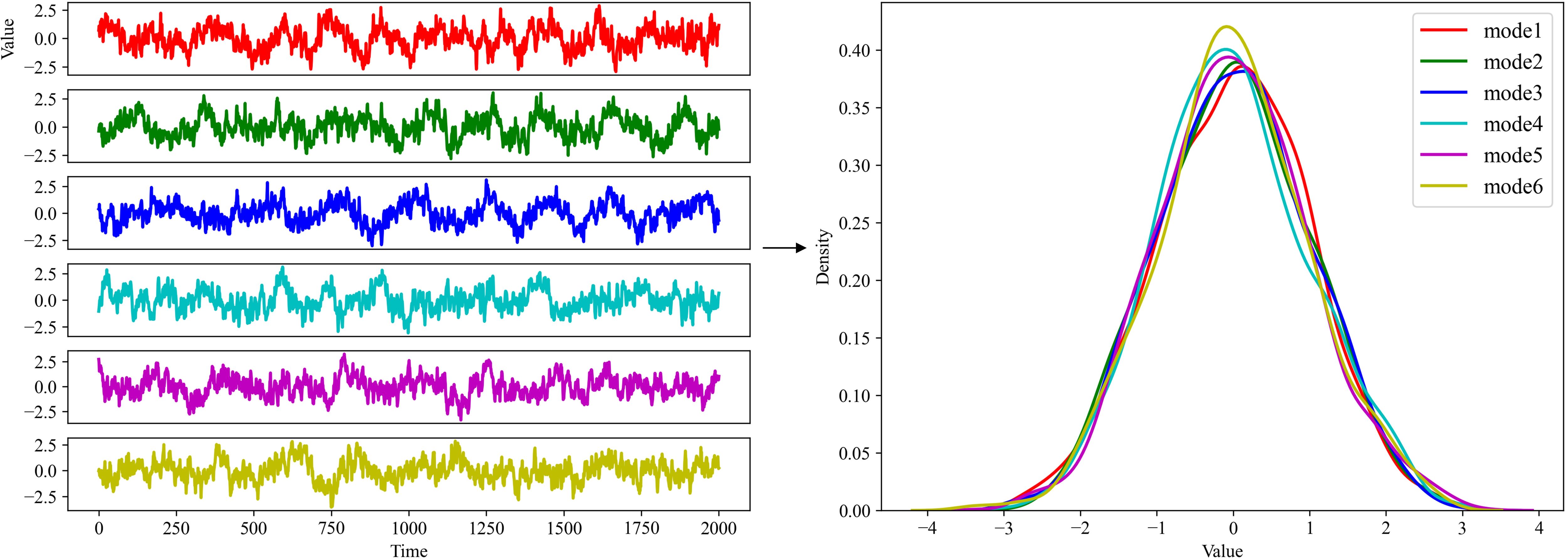}
		\caption{{Normal data.}}
	\end{subfigure}
	\begin{subfigure}[b]{1\linewidth}
		\centering
		\includegraphics[width=1.\textwidth,height=0.3574\textwidth]{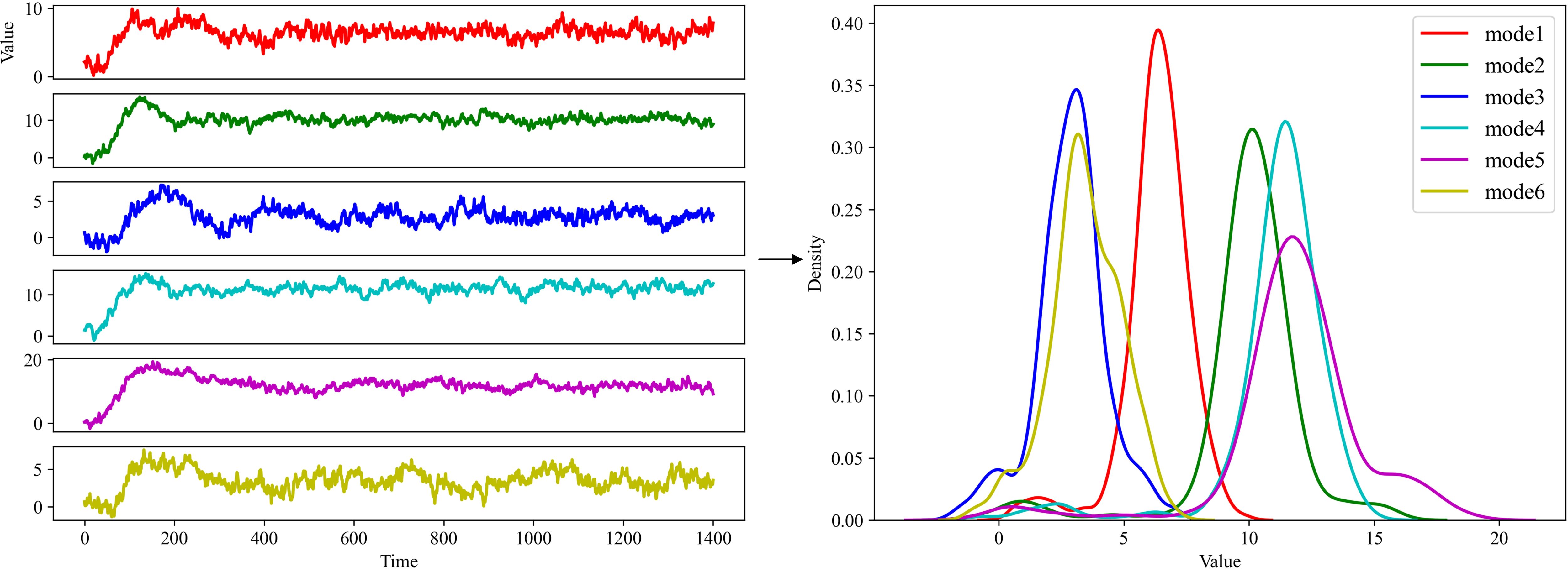}
		\caption{{Fault 2 data.}}
	\end{subfigure}
	\caption{Normal and fault 2 data of material A flow rate after standardization of the Tennessee Eastman process in six modes.}
 \label{Fig1}    
\end{figure}

\hyperref[Fig1]{Fig. 1} shows the normal data and fault 2 data of material A flow rate after standardization of the Tennessee Eastman process in six modes, respectively. The left shows the curve of the variable over time, and the right represents the probability density function of the variable. It can be seen that the distribution difference of normal data after standardization can be ignored, while the distribution difference of fault 2 data still exists. 
For clarity, the following assumptions are made:
The features contained in the standardized monitoring data can be divided into domain-invariant and domain-specific features. 
Domain-invariant features remain consistent across different modes, while domain-specific features vary with modes. Domain-invariant features can be used to identify the system health condition class, but not mode class. While domain-specific features can be taken to determine both system health condition class and mode class.
As can be seen from \hyperref[Fig1]{Fig. 1}, the data distribution of material A flow rate shifts in the positive direction, which can be regarded as a domain-invariant feature of the fault 2. And the magnitude of the increase in the flow rate of material A or the specific details of the change can be regarded as a domain-specific feature of fault 2. Both domain-invariant and domain-specific features can distinguish the system health condition class under a certain mode. The key to single-domain generalization in multi-mode fault diagnosis is to extract domain-invariant features.

With the aim of generalizing the fault diagnosis model constructed on single-mode dataset to other modes, a feasible idea is to generate diverse samples, which should cover sample space of unseen modes as much as possible. From this perspective, the generalization performance of the fault diagnosis model can be limited by the generated data. However, directly generating samples of unseen modes is complex. Therefore, generating samples of unseen modes is transformed into generating pseudo-sample features. 
{Based on the assumption that samples with the same health condition category across different modes share domain-invariant features, the sample features of the unseen modes can be generated based on those of the single seen mode through specific mapping relationships. These features are referred to as pseudo-sample features. 
Pseudo-sample features are crucial in the single-domain generalization task, as they allow the fault diagnosis model to explicitly access fake unseen modes that have significant distribution differences from the single seen mode. 
Specifically, these features introduce variability to simulate potential differences between the seen and unseen modes. This allows the fault diagnosis model to generalize beyond the scope of the single seen mode. 
Pseudo-sample features provide the fault diagnosis model with a mechanism to learn cross-domain features, thereby enhancing its ability to handle the target domain with unforeseen distribution shifts.} 
Additionally, in the process of generating pseudo-sample features, how to retain domain-invariant features while increasing diversity needs to be focused on. 

\subsection{Diversity of generated samples}
The purpose of increasing sample diversity is to ensure that the generated samples cover as much of the unseen mode sample space as possible, thereby enhancing the completeness of the training samples.  The upper bound of mutual information was minimized to encourage diversity distributions in \cite{RN187888}. The multi-scale style generation strategy and adversarial contrastive learning strategy are adopted to enable the generated samples more diverse in \cite{RN186301}. An incremental domain augmentation strategy is proposed to generate augmented domains with difference and completeness in \cite{RN237026}.  Multi-scale deep separable convolutional kernels and Mix-up are combined to generate extended domain samples with different distributions in \cite{RN237028}. These methods achieve the generation of diverse samples. During the sample generation process, domain-invariant features remain unchanged. Therefore, sample generation should be constrained by domain-invariant features. 

\subsection{Domain-invariant feature extraction}
Domain-invariant features do not change with different modes. Therefore, in single-source domain generalization for fault diagnosis, extracting domain-invariant features helps inference the system health state across domains. Lower bound of mutual information was maximized to learn semantic-consistent representations in \cite{RN187888}. 
Contrastive learning is used to learn class-wise domain-invariant representations in \cite{RN186301,RN237026}. Adversarial training and metric learning strategy is designed to learn generalized features in \cite{RN237028}. For samples with strong diversity, supervised contrastive learning and adversarial learning alone may not be able to learn domain-invariant features well. Therefore, different strategies need to be combined to enhance the learning of domain-invariant features.

\section{Proposed method}
\label{Sec3}
\subsection{Overall structure of DACN}

\begin{figure}[!ht]
    \centering
    \includegraphics[width=1.\textwidth,height=0.52215\textwidth]{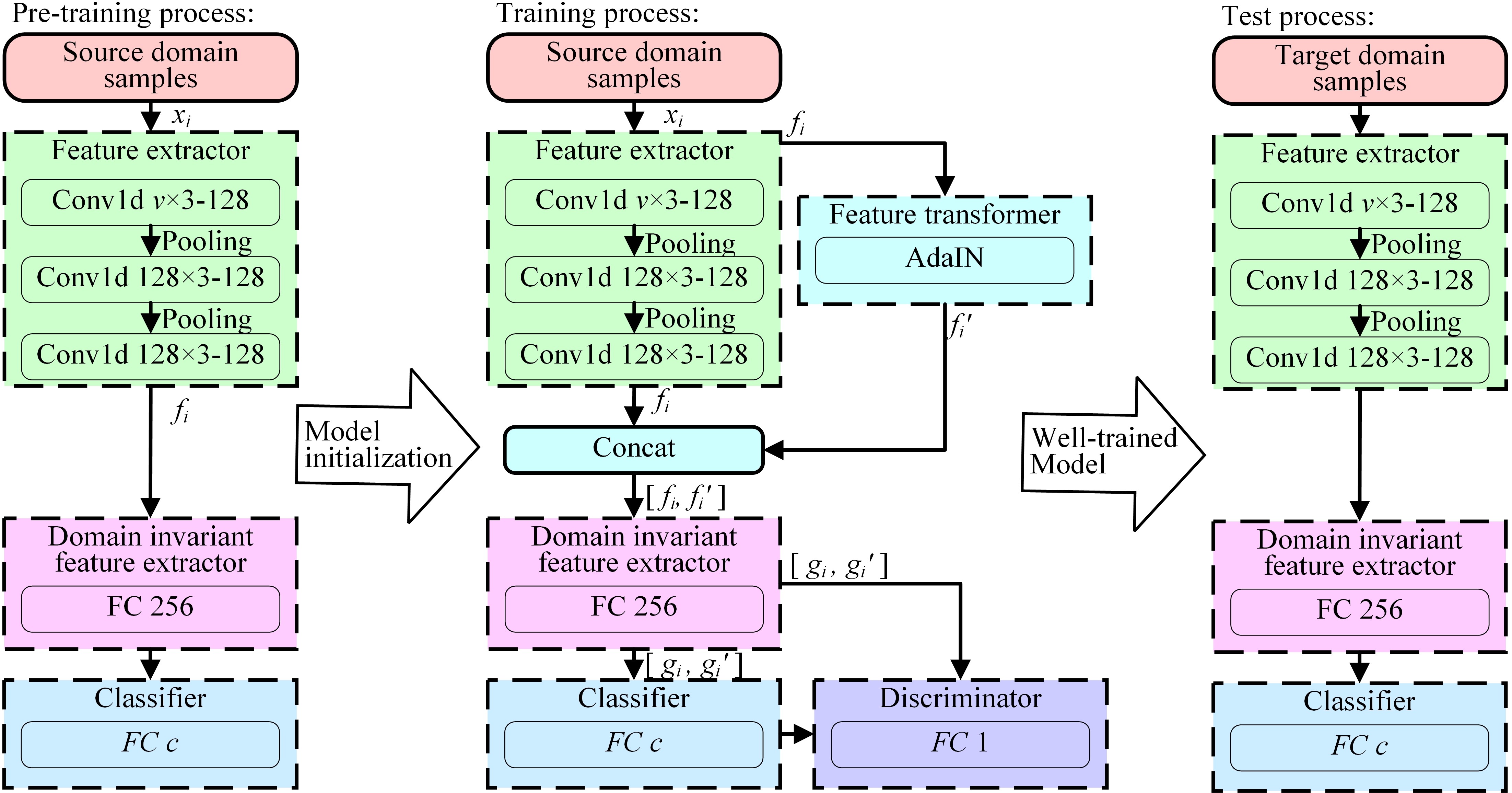}
    \caption{Overall structure of DACN.}
    \label{Fig2}
\end{figure}

DACN is trained with samples from single seen mode and tested with samples from unseen modes. The overall structure of DACN is shown in \hyperref[Fig2]{Fig. 2}.
Unlike other single-source domain generalization fault diagnosis methods, such as AMINet\cite{RN187888} and MSG-ACN\cite{RN186301},  the training of DACN involves two steps: pre-training process and training process. This approach helps prevent the generator from being guided by a poorly optimized classifier, thereby reducing the risk of generating incorrect samples. 
Both pre-training process and training-process are based on a single-mode samples for training. After the pre-training process, DACN would capture the sample features of the single seen mode related to the system health condition. Then during the training process, DACN would generate diverse pseudo-sample features based on seen mode sample features, extract domain-invariant features, and then determine the system health condition classes. After these two processes, DACN is well-trained. DACN consists of data processing, feature extractor $F$, feature transformer $H$, domain invariant feature extractor $G$, classifier $C$ and discriminator $D$. The detailed composition of each module is shown in \hyperref[Table1]{Table 1}, where $v$ is the number of measured variables.

\begin{table}[!ht]

\caption{Structure of the modules in DACN.}
\begin{tabular}{lll}
\hline
Modules                            & Type            & (Input size, output size, kernel size) \\ \hline
\multirow{5}{*}{Feature extractor} & Conv1d          & ($v$×64, 128×64, $v$×3)                   \\
                                   & Max pooling     & (128×64, 128×32, -)                   \\
                                   & Conv1d          & (128×32, 128×32, 128×3)               \\
                                   & Max pooling     & (128×32, 128×16, -)                   \\
                                   & Conv1d          & (128×16, 128×16, 128×3)               \\
Feature transformer          & -               & (128×16, 128×16, -)                   \\
Domain invariant feature extractor & Fully connected & (2048,256, -)                         \\
Classifier                         & Fully connected & (256, c, -)                           \\
Discriminator                      & Fully connected & ((256×c), 1, -)                       \\ \hline
\end{tabular}
\label{Table1}
\end{table}

\subsubsection{Data processing}
Data processing mainly consists of standardization and time series expansion. As mentioned in \hyperref[Sec2-2]{section \ref{Sec2-2}}, it is assumed that the mode features can be represented by the mean the fault-free data. Thus, standardization is performed to remove mode features in the data. Time-series monitoring data contains rich spatiotemporal feature information. Time series expansion is employed to preserve the temporal features of the data. Time series expansion is implemented by expanding the data of the previous $k-1$ moments to the current moment.
\subsubsection{Feature extractor}
Feature extractor $F$ takes the samples after data processing as inputs and outputs deep-level features. It maps the sample space to the feature space using three 1-dimensional convolutional layers and two max-pooling layers. ReLU is used as the activation function \cite{RN237029}. The activation function is located after each maximum pooling layer and the last 1-dimensional convolutional layer. Let $f_{i}$ denotes the feature representations of sample $x_i$, 	$f_i=F\left(x_i\right)$.
\subsubsection{Feature transformer}

In image style transfer, AdaIN achieves style learning by aligning the mean and variance of features between two different style images, such as handwritten and printed digits. Similarly, in multi-mode fault diagnosis, AdaIN can facilitate mode learning by aligning the mean and variance of sample features from different modes. This approach typically requires training the model with samples from at least two modes to effectively capture the variations between them. 
However, in single-source domain generalization for fault diagnosis, only samples from a single domain are available, making it impossible to learn specific mode variations. 

The deep-level features extracted by feature extractor contains domain-invariant and domain-specific features. From \hyperref[Fig1]{Fig. 1}, it can be seen that the change trend and change amplitude of the fault-related variables can be regarded as domain-invariant and domain-specific features of fault 2, respectively. Therefore, unseen mode sample features can be simulated by varying the magnitude of single seen mode sample features. 
Feature transformer $H$ generates pseudo-sample features $f_i’$ based on the single seen mode sample features $f_i$. $f_i’=H\left(f_i,n_1,n_2\right)={h_1\left(n_1\right)\left(f_i-\mu_i\right)}/{\sigma_i}+h_2\left(n_2\right)$, where $n_1$ and $n_2$ are random noise, $n_1$ is set to be sampled from a uniform distribution, and $n_2$ is set to be sampled from a standard normal distribution; $h_1$ and $h_2$ are fully connected layers; $\mu_i$ and $\sigma_i$ is the mean and variance of $f_i$. $h_1\left(n_1\right)$ and $h_2\left(n_2\right)$ are used to simulate the variations in feature magnitude and steady-state bias across different modes. 
In \cite{RN255602}, $n_1$ was set as random noise sampled from a standard normal distribution $n_1 \sim N(0,1)$ , which results in sampled values concentrated around the mean of 0. Consequently, the output values $h_1\left(n_1\right)$ would exhibit minimal variation, making it difficult to model significant differences in sample features. Furthermore, the magnitude of sample features in unseen modes may be either amplified or reduced compared to the single seen mode. Since the magnitude variation is determined by $h_1\left(n_1\right)$, the specific range of magnitude scaling cannot be precisely defined. Therefore, $n_1$ is set as $n_1 \sim U(0.05, 1.95)$ to introduce a broader range of sampled values compared to a normal distribution centered around 0. This increases the diversity of sampled values and indirectly influences the magnitude scaling, enhancing the model's ability to simulate the variations in unseen mode sample features.

Concat is the module to splice single seen mode sample features and pseudo-sample features in the batch dimension.

\subsubsection{Domain-invariant feature extractor}
Both single seen mode sample features and pseudo-sample features contain domain-invariant and domain-specific features. And only domain-invariant features contribute to multi-mode fault diagnosis. The domain-invariant feature extractor $G$ is expected to implement the function of extracting domain-invariant features from single seen mode sample features and pseudo-sample features through a fully connected layer. The activation
function is ReLU. Dropout is added to avoid overfitting and enhance the robustness of the model. It allows the model to effectively identify system health condition even when there are slight differences in the sample features (i.e., the specific measured variables affected by a fault and the trends in these variable changes) across different modes. Let $g_i$ and $g_i'$ denote the domain-invariant features of single seen mode data and pseudo-sample data, respectively. $g_i$ and $g_i'$ are defined as, $\begin{bmatrix}g_i,g_i'\end{bmatrix}=G(\begin{bmatrix}f_i,f_i'\end{bmatrix})$. 

\subsubsection{Classifier}

The classifier $C$ takes the extracted domain-invariant features as inputs and outputs the system health condition class. It is also implemented through a fully connected layer. Let $c_i$ and $c_i'$ denote the classifier predictions of single seen mode data and pseudo-sample data, respectively. $c_i$ and $c_i'$ are defined as, $\begin{bmatrix}c_i,c_i'\end{bmatrix}=C(\begin{bmatrix}g_i,g_i'\end{bmatrix})$. 
\subsubsection{Discriminator}
Mode classes are distinguished by 
Discriminator $D$. In multi-mode fault diagnosis, monitoring data exhibits multi-modal structural characteristics. Long et al. proposed that the multimodal structures can only be captured sufficiently by the cross-covariance dependency between the features and classes \cite{RN237019}. Thus, the discriminator is conditioned on the cross-covariance of domain-invariant feature representations and classifier predictions. The discriminator predictions of single seen mode data and pseudo-sample data are defined as, $\begin{bmatrix}d_i,d'_i\end{bmatrix}=D\left(\begin{bmatrix}g_i{\otimes}c_i,g'_i{\otimes}c'_i\end{bmatrix}\right)$, where $\otimes$ is a multilinear mapping.

\subsection{Pre-training process}
The pre-training process enables the DACN model to learn single seen mode sample features in advance, facilitating the generation of more effective pseudo-sample features during later training process. The fault diagnosis model is established on the single seen mode monitoring data. Thus, it should first have high classification accuracy on this data set. The fault diagnosis model is optimized through cross-entropy loss of single seen mode data, which is defined as,

\begin{equation}
    L_{\mathrm{c}1}=-\frac{1}{N_{\mathrm{s}}}\sum_{i=1}^{N_{\mathrm{s}}}\sum_{l=1}^{L}y_{i}^{l}\log c_{i}^{l}
    \label{Eq1}
\end{equation}
where $y_{i}^{l}$ is the $l$-th dimension of the true label of the $i$-th sample, and $c_{i}^{l}$ is the $l$-th dimension of the classifier prediction of the $i$-th sample.
\subsection{Training process}
During training process, 
DACN should accurately classify single seen mode samples and ensure pseudo-samples have sufficient semantic information and diversity, enabling domain-invariant feature extraction. To ensure the accurate classification of single seen mode samples, \hyperref[Eq1]{Eq. (1)} is used to optimize the fault diagnosis performance of the model.

\subsubsection{Semantic preservation of pseudo-sample features}
The single seen mode sample features and pseudo-sample features serve as the inputs and outputs of the feature transformer, respectively. They possess different domain-specific features but share the same domain-invariant features, meaning that their system health condition classes are identical. The system health condition class labels of the single seen mode data are assigned to the corresponding pseudo-sample features. 
The label classification loss for the pseudo-sample features is defined as,

\begin{equation}
    L_{\mathrm{c}2}=-\frac{1}{N_{\mathrm{s}}}\sum_{i=1}^{N_{\mathrm{s}}}\sum_{l=1}^{L}y_{i}^{l}\log {{c'}_{i}^{l}}
    \label{Eq2}
\end{equation}
where ${c'}_{i}^{l}$ is the $l$-th dimension of the classifier prediction of the $i$-th pseudo-sample features.

\subsubsection{Domain invariant feature extraction}
\label{Sec3-3-2}
Sample features extracted from single mode monitoring data are difficult to discern whether they are domain-invariant or domain-specific features. domain-invariant features need to be extracted from multi-mode monitoring data. pseudo-sample features represent the sample features of unseen modes.
Thus, domain-invariant features can be extracted from single seen mode sample features and pseudo-sample features. 

The domain-invariant feature extraction aims to enhance the consistency of sample features across different modes and reduce the discrepancies caused by mode variations. Therefore, the domain-invariant feature extractor needs to learn a mapping that minimizes the differences in sample features across different modes.
Supervised contrastive learning can bring samples belonging to the same class closer in the embedding space while separating samples from different classes \cite{RN237018}. It is used here to pull together samples belonging to the same class in the embedding space , thereby reducing the feature differences of the same system health condition class across different modes. The supervised contrastive loss is defined as,

\begin{equation}
    \begin{aligned}
&L_{\mathrm{sup}} =\sum_{i=1}^{2N_\mathrm{s}}L_i^\mathrm{sup},
L_{i}^{\sup} =\frac{-1}{2N_\mathrm{S}-1}\sum_{j=1}^{2N_\mathrm{S}}1_{i\neq j}\cdot1_{y_i=y_j}\cdot\log\frac{\exp\left(g_i\cdot g_j / \tau\right)}{\sum\limits_{k=1}^{2N_\mathrm{s}}1_{i\neq k}\cdot\exp\left(g_i\cdot g_k / \tau\right)} 
\end{aligned}
    \label{Eq3}
\end{equation}
where $1_{A}$ is an indicator function, which means it is 1 when $A$ is satisfied, otherwise it is 0. $\tau$ is a scalar temperature parameter. Since the single seen mode sample features $g$ and pseudo-sample features $g’$ are mixed together, they are uniformly recorded as $g$ in \hyperref[Eq3]{Eq. (3)}.

Domain-invariant features may not be effectively extracted using only supervised contrastive learning, especially between samples with significant mode differences. 
Supervised contrastive learning aids in domain-invariant feature extraction by reducing the feature differences of the same system health condition class across different modes to a certain extent. However, based on the assumption of domain-invariant features, they can be used to identify system health condition classes but are unable to distinguish mode classes. To achieve this functionality, adversarial learning is employed.
In adversarial learning, domain-invariant features are extracted through adversarial training of domain-invariant feature extractor $G$ and discriminator $d$.  Maximizing and minimizing the discriminative loss are the optimization objectives for the feature extractor and discriminator, respectively. The discriminative loss is defined as,

\begin{equation}
    L_\mathrm{d}=-\sum_{i=1}^{N_\mathrm{S}}\log d_i-\sum_{i=1}^{N_\mathrm{S}}\log\left(1-d_i'\right)
    \label{Eq4}
\end{equation}

The gradient reversal layer(GRL) between domain-invariant feature extractor and discriminator implements adversarial training, as shown in \hyperref[Fig3]{Fig. 3}. The adversarial training is described as follows: 

The mode class of the sample is identified by the discriminator, determining whether the current sample belongs to the single seen mode.
By minimizing the discriminative loss, discriminator learns the differences in modes as much as possible. Thus, discriminator can be considered as an expert of domain-specific features. If domain-specific features exist, discriminator can capture them.

The optimization objective of the domain invariant feature extractor is to maximize the discriminative loss. The domain invariant feature extractor reduces the differences between different modes in embedding space. This means that the sample features extracted by the domain-invariant feature extractor cannot contain domain-specific features. By the adversarial training, domain-invariant feature extractor and discriminator become experts of domain-invariant and domain-specific features, respectively.

\subsubsection{Diversity of pseudo-sample features}

The analysis in  \hyperref[Sec3-3-2]{section \ref{Sec3-3-2}} shows that the domain-invariant feature extractor can extract domain-invariant features, and the discriminator can capture domain-specific features. But the premise is that the input of the domain invariant feature extractor is multi-mode monitoring data. 

\begin{figure}[!ht]
    \centering
    \includegraphics[width=1.\textwidth,height=0.51468\textwidth]{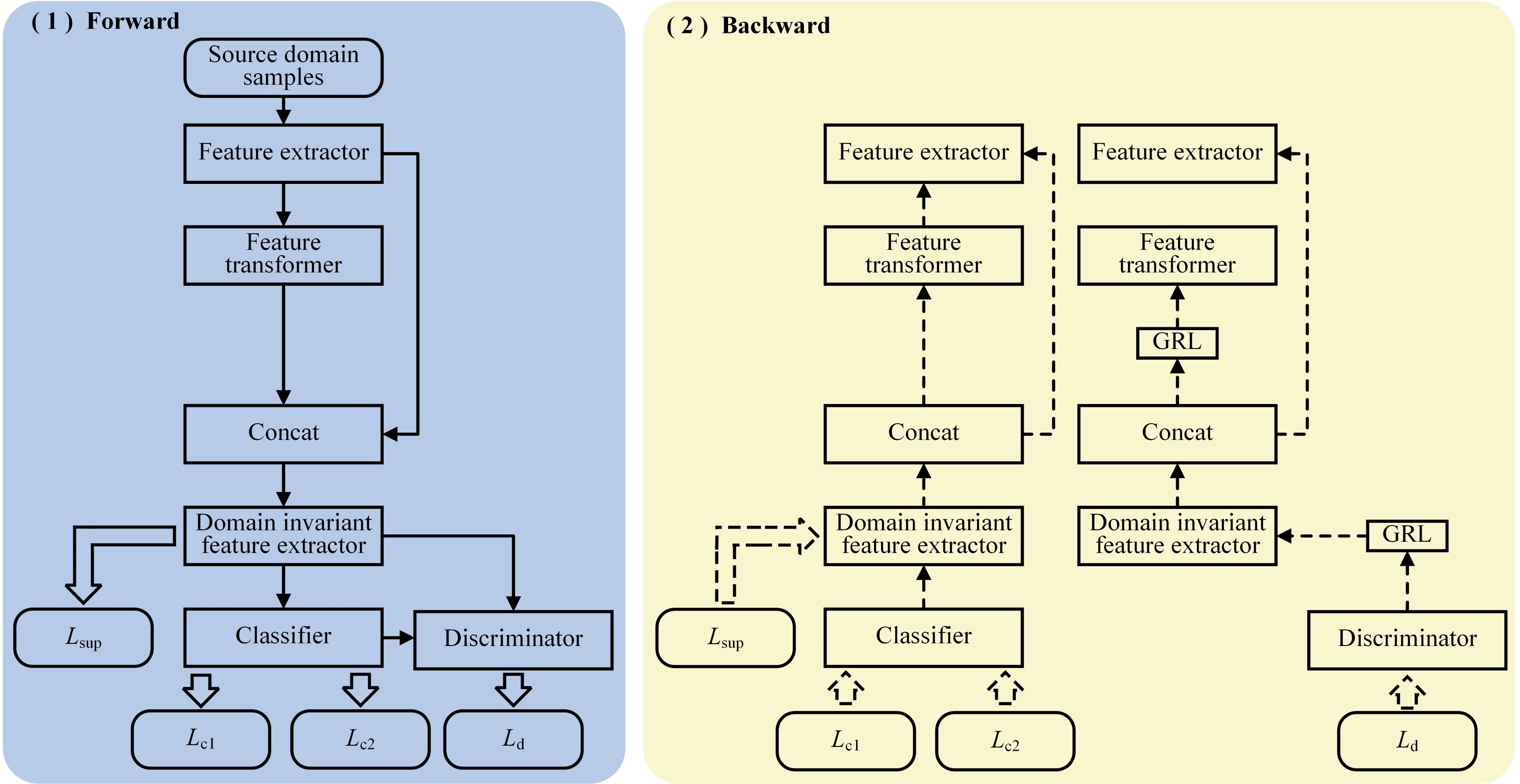}
    \caption{Forward and backward propagation of DACN.}
    \label{Fig3}
\end{figure}

By maximizing the discriminative loss, the domain-invariant feature extractor can reduce the difference between single seen mode sample features and pseudo-sample features in the embedding space. If the optimization objective of the feature transformer is also to maximize the discriminative loss, then the difference between its generated pseudo-sample features and the single seen mode sample features will be reduced. The generated pseudo-sample features and single seen mode sample features cannot be considered to belong to different modes. This contradicts the premise of \hyperref[Sec3-3-2]{section \ref{Sec3-3-2}}. Thus, the optimization objective of the feature transformer is set to minimize the discriminative loss, resulting in a large difference between the generated pseudo-sample features and the single seen mode sample features. There is also a GRL between feature transformer and domain-invariant feature extractor, as shown in \hyperref[Fig3]{Fig. 3}. 

\begin{figure}[!ht]
    \centering
\includegraphics[width=1.\textwidth,height=0.316\textwidth]{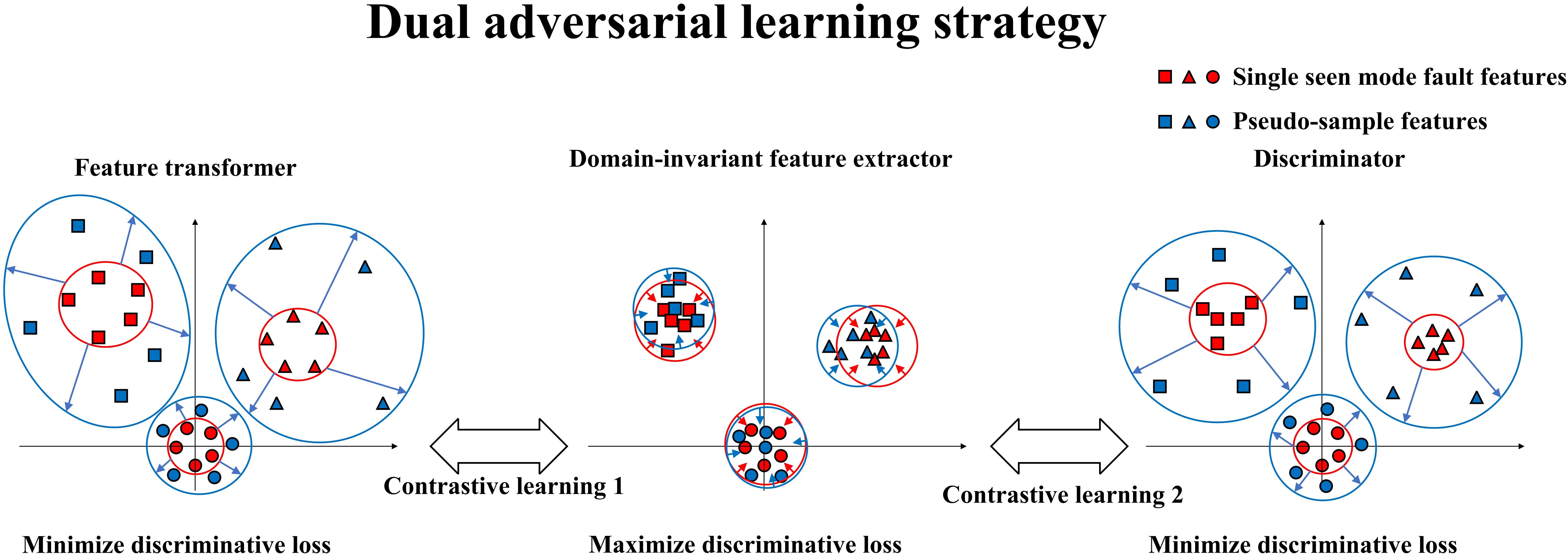}
    \caption{Two sets of adversarial learning in DACN.}
    \label{Fig4}
\end{figure}

The adversarial training between the domain-invariant feature extractor and the feature transformer is beneficial for diagnosing system health condition classes in unseen mode samples. The domain-invariant feature extractor reduces the differences between single seen mode sample features and pseudo-sample features in the embedding space by maximizing the discriminative loss. However, the feature transformer increases the difference between single seen mode sample features and pseudo-sample features by minimizing the discrimination loss. This process poses a challenge for the domain-invariant feature extractor, requiring it to embed highly differentiated sample features into a less differentiated embedding space. During this process, the pseudo-sample features simulate challenging samples with difficult-to-extract domain-invariant features, which helps the model adapt to scenarios with significant mode differences. Overall, DACN implements two sets of adversarial learning, as illustrated in \hyperref[Fig4]{Fig. 4}. The first set involves domain-invariant feature extractor and discriminator, encouraging the domain-invariant feature extractor to learn domain-invariant features. The second set involves domain-invariant feature extractor and feature transformer, enhancing the feature transformer to generate more diverse pseudo-sample features.

\subsubsection{The overall optimization objectives of DACN}

To obtain a unified loss function, the maximization of discriminative loss is converted into the minimization of the negative discriminative loss. The overall loss function used to  optimize feature extractor, domain-invariant feature extractor and classifier is defined as,

\begin{equation}
    L_1=\lambda_1L_{\mathrm{c}1}+\lambda_2L_{\mathrm{c}2}+\lambda_3L_{\sup}-\lambda_4L_{\mathrm{d}}
    \label{Eq5}
\end{equation}
where $\lambda_1$, $\lambda_2$, $\lambda_3$ and $\lambda_4$ are hyperparameters.

The loss function of discriminator is defined as,

\begin{equation}
    L_2=L_\mathrm{d}
    \label{Eq6}
\end{equation}

The loss function of feature transformer is defined as,

\begin{equation}
    L_3=\lambda_1L_{\mathrm{c}2}+\lambda_3L_{\sup}+\lambda_4L_{\mathrm{d}}
    \label{Eq7}
\end{equation}

The training and inference of DACN is shown in \hyperref[alg1]{Algorithm 1}.

\begin{algorithm}[!ht]  
	\caption{DACN.}  
	\label{alg1}  
\# \textbf{Pre-training stage}\\
\textbf{Input: }Source domain dataset $D_S$.\\
\textbf{Model: }feature extractor $F$, domain-invariant feature extractor $G$ and classifier $C$.\\
\textbf{Output: }feature extractor $F$, domain-invariant feature extractor $G$ and classifier $C$.\\
1: Perform data processing on the source domain dataset \\
2: \textbf{for} \textit{epoch} = 1 to \textit{epochs} \textbf{do}\\
3:\ \ \ \ \ \textbf{for} \textit{batch} = 1 to \textit{batches} \textbf{do}\\
4:\ \ \ \ \ \ \ \ \ Calculate classifier output.\\
5:\ \ \ \ \ \ \ \ \ Calculate loss $L_{\mathrm{c}1}$ via \hyperref[Eq1]{Eq. (1)}.\\
6:\ \ \ \ \ \ \ \ \ Update parameters of the model through gradient descent algorithm.\\
7:\ \ \ \ \ \textbf{end for}\\
8: \textbf{end for}\\
\textbf{Return:} Pre-trained model.\\
\\
\# \textbf{Training stage}\\
\textbf{Input: }Source domain dataset $D_S$ and pre-trained model.\\
\textbf{Model: }feature extractor $F$, feature transformer $H$, domain-invariant feature extractor $G$, classifier $C$ and discriminator $D$.\\
\textbf{Output: }feature extractor $F$, domain-invariant feature extractor $G$, and classifier $C$.\\
1: Perform data processing on the source domain dataset.\\
2: Initialize feature extractor $F$, domain-invariant feature extractor $G$ and classifier $C$ using parameters obtained from pre-training.\\
3: \textbf{for} \textit{epoch} = 1 to \textit{epochs} \textbf{do}\\
4:\ \ \ \ \ \textbf{for} \textit{batch} = 1 to \textit{batches} \textbf{do}\\
5:\ \ \ \ \ \ \ \ \ Calculate classifier and discriminator output.\\
6:\ \ \ \ \ \ \ \ \ Calculate the losses $L_1$, $L_2$ and $L_3$ via \hyperref[Eq5]{Eq. (5)} to \hyperref[Eq7]{Eq. (7)}.\\
7:\ \ \ \ \ \ \ \ \ Update parameters of the model through gradient descent algorithm.\\
8:\ \ \ \ \ \textbf{end for}\\
9: \textbf{end for}\\
\textbf{Return:} feature extractor $F$, domain-invariant feature extractor $G$ and classifier $C$\\
\\
\# \textbf{Inference stage}\\
\textbf{Input: }Target domain dataset $D_T$.\\
\textbf{Model: }feature extractor $F$, domain-invariant feature extractor $G$ and classifier $C$.\\
\textbf{Output: }System health condition class.
\end{algorithm}  

\section{Experiments}
\label{Sec4}
Tennessee Eastman (TE) process \cite{RN237020} and CSTR \cite{RN237031} are used for single-source domain generalization in fault diagnosis experiments.
\subsection{TE}
\begin{figure}[!ht]
    \centering
    \includegraphics[width=1.\textwidth,height=0.5632\textwidth]{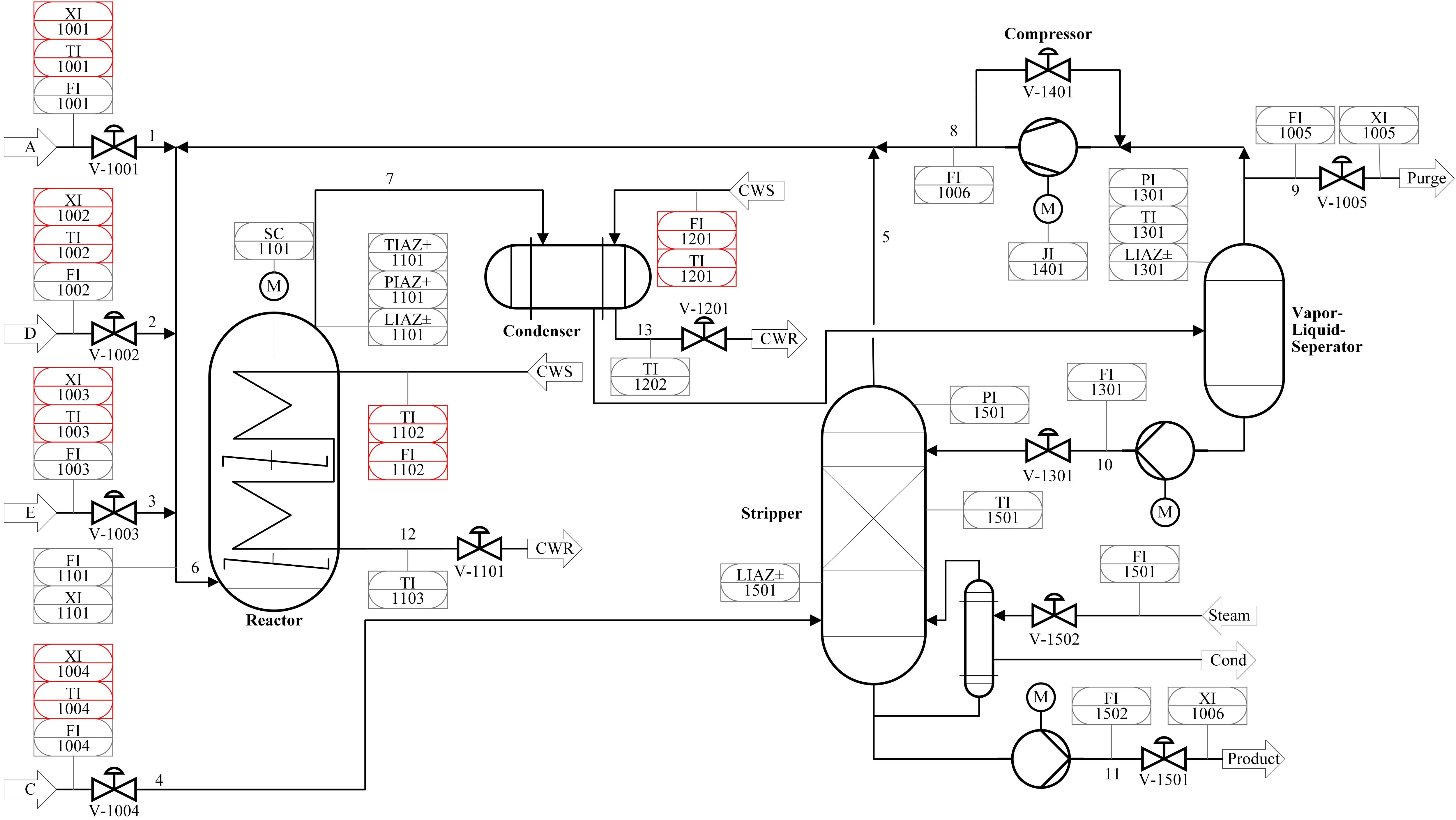}
    \caption{P\&ID of the revised process model \cite{RN237021}.}
    \label{Fig5}
\end{figure}

TE process has become a benchmark for evaluating process monitoring and fault diagnosis models. Its simulation model has been implemented through SIMULINK \cite{RN237021}. Liu et al. obtained models for six modes by adjusting the parameters \cite{RN237030}. The structure of the TE process is shown in \hyperref[Fig5]{Fig. 5}. The process consists of 53 measured variables and can simulate 28 faults in 6 different modes. The fault, mode and task settings used to validate the model are presented in \hyperref[Table2]{Table 2} - \hyperref[Table4]{Table 4}, respectively.

\begin{table}[!ht]
\centering
\caption{Faults of TE process used in single source domain generalization fault diagnosis \cite{RN237020}.}
\label{Table2}
\begin{tabular}{lll}
\hline
Fault No. & Description                               & Type             \\ \hline
F0        & Normal                                    & -                \\
F2        & B composition, A/C ratio constant         & Step             \\
F4        & Reactor cooling water inlet temperature   & Step             \\
F8        & A, B, C feed composition                  & Random variation \\
F10       & C feed temperature                        & Random variation \\
F11       & Reactor cooling water inlet temperature   & Random variation \\
F12       & Condenser cooling water inlet temperature & Random variation \\
F13       & Reaction kinetics                         & Slow drift       \\
F14       & Reactor cooling water valve               & Sticking         \\
F17       & Unknown                                   & -                \\ \hline
\end{tabular}
\end{table}

\begin{table}[!ht]
\centering
\caption{Modes of TE process used in single source domain generalization fault diagnosis \cite{RN237020}.}
\label{Table3}
\begin{tabular}{lll}
\hline
Mode No. & G/H mass ratio & Production rate           \\ \hline
M1       & 50/50          & G: 7038kg/h, H: 7038kg/h  \\
M2       & 10/90          & G: 1408kg/h, H: 12669kg/h \\
M3       & 90/10          & G: 10000kg/h, H: 1111kg/h \\
M4       & 50/50          & maximum production rate   \\
M5       & 10/90          & maximum production rate   \\
M6       & 90/10          & maximum production rate   \\ \hline
\end{tabular}
\end{table}

\begin{table}[!ht]
\centering
\caption{Single source domain generalization fault diagnosis task description of TE process.}
\label{Table4}
\begin{tabular}{llll}
\hline
\begin{tabular}[c]{@{}l@{}}Task \\ No.\end{tabular} & \begin{tabular}[c]{@{}l@{}}Training set (No. of \\ samples: 11200)\end{tabular} & \begin{tabular}[c]{@{}l@{}}Test set 1 (No. of \\ samples: 2810)\end{tabular} & \begin{tabular}[c]{@{}l@{}}Test set 2 (No. of \\ samples: 70050)\end{tabular} \\ \hline
T1                                                  & M1                                                                              & M1                                                                           & M2, M3, M4, M5 and M6                                                         \\
T2                                                  & M2                                                                              & M2                                                                           & M1, M3, M4, M5 and M6                                                         \\
T3                                                  & M3                                                                              & M3                                                                           & M1, M2, M4, M5 and M6                                                         \\
T4                                                  & M4                                                                              & M4                                                                           & M1, M2, M3, M5 and M6                                                         \\
T5                                                  & M5                                                                              & M5                                                                           & M1, M2, M3, M4 and M6                                                         \\
T6                                                  & M6                                                                              & M6                                                                           & M1, M2, M3, M4 and M5                                                         \\ \hline
\end{tabular}
\end{table}

In the process of generating datasets through SIMULINK simulation, the sampling interval is set to 3 seconds, and the simulation time is set to 100 hours, with faults implanted starting from 30 hours. Each class will produce 1400 samples in each mode. The length of the time series expansion is set to 64. Data from each mode is sequentially chosen as source domain dataset, and data from the other five modes served as the target domain dataset (test set 2). The source domain dataset is divided into training set and test set 1. The training set and test set 1 are randomly divided, with proportions of 80\% and 20\%, respectively. The training set is used to build a fault diagnosis model. Test set 1 is used to determine if the model has basic diagnostic capabilities on data distributions similar to the training data. Test set 2, which is the focus of comparison, is used to evaluate the model's diagnostic performance on unseen modes. By comparing the model's performance on test set 1 and test set 2, the model learns whether more domain-invariant features or domain-specific features can be characterized.

\subsection{CSTR}

\begin{figure}[!ht]
    \centering
    \includegraphics[width=0.5\textwidth,height=0.2244\textwidth]{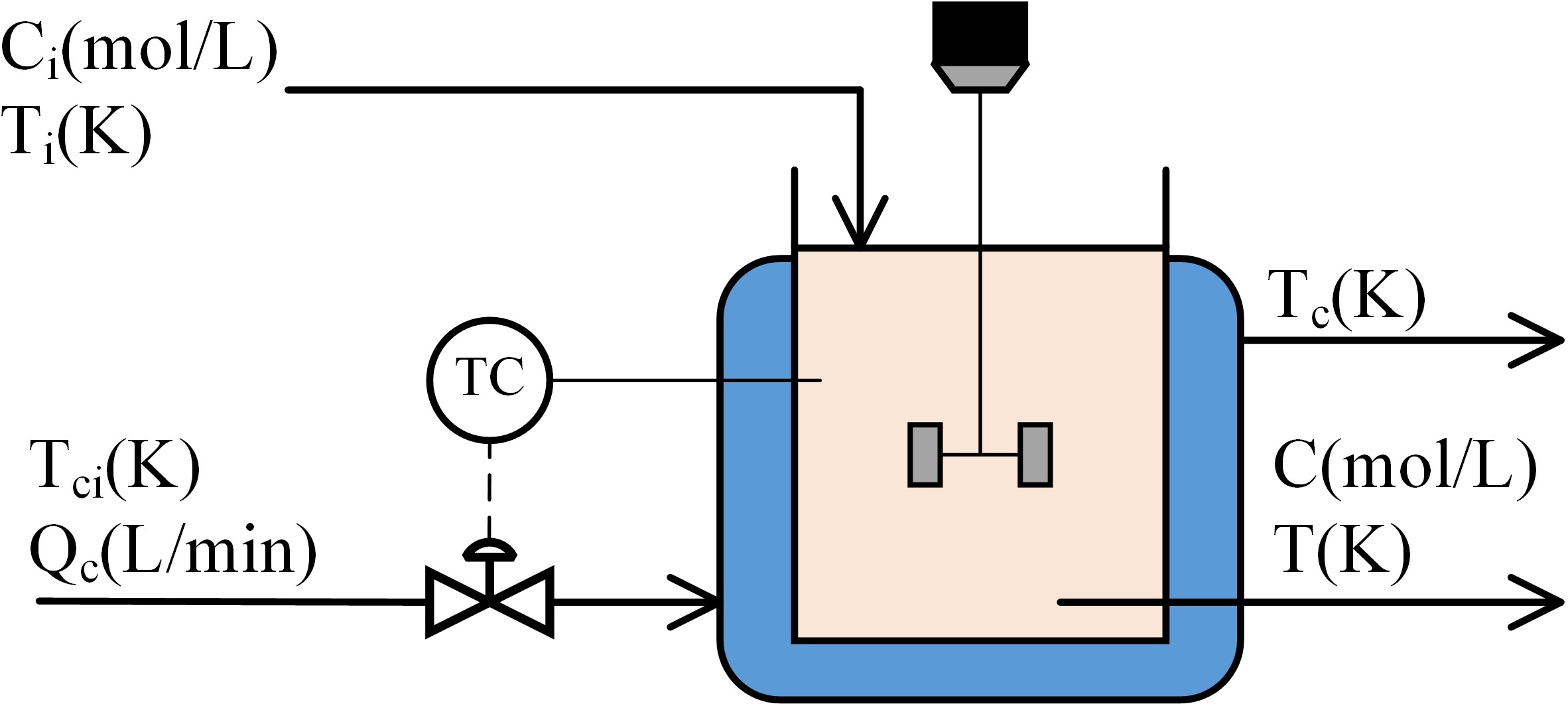}
    \caption{Structure of closed-loop CSTR \cite{RN237031}.}
    \label{Fig6}
\end{figure}

Closed-loop CSTR has been also widely used to validate the performance of fault diagnosis algorithms. The measured variables of the process are shown in \hyperref[Fig6]{Fig. 6}. The reactor temperature (T) is controlled by the volumetric flow rate of the cooling water (Q$_\mathrm{c}$).

This process can be described by follows,

\begin{equation}
    \begin{aligned}
&\frac{dC}{dt}=\frac{Q}{V}\Big(C_{i}-C\Big)-kC \\
&\frac{dT}{dt}=\frac{Q}{V}\Big(T_{i}-T\Big)-\frac{(\Delta H_{r})kC}{\rho C_{p}}-\frac{UA}{\rho C_{p}V}\Big(T-T_{c}\Big) \\
&\frac{dT_{c}}{dt}=\frac{Q_{c}}{V_{c}}\Big(T_{ci}-T_{c}\Big)+\frac{UA}{\rho_{c}C_{pc}V_{c}}\Big(T-T_{c}\Big) \\
&k=k_{0}\exp\biggl(\frac{-E}{RT}\biggr)
\end{aligned}
    \label{Eq8}
\end{equation}
where the detailed parameter information can be seen in \cite{RN237031}. The faults set in CSTR are shown in \hyperref[Table5]{Table 5}. Data for mode 1 (M1) were collected at the original set point. Data for mode 2 (M2) and mode 3 (M3) were collected with set points increased by 5K and 10K, respectively. The simulation time was set to 20 hours, with faults introduced at 200 minutes. The division of the dataset followed the same experimental setup as the TE process.  Detailed task settings are shown in \hyperref[Table6]{Table 6}.

\begin{table}[!ht]
\centering
\caption{Faults of CSTR used in single source domain generalization fault diagnosis.}
\label{Table5}
\begin{tabular}{ll}
\hline
Fault No. & Description                         \\ \hline
F0        & Normal                              \\
F1        & Reactant inlet concentration        \\
F2        & Reactant inlet temperature          \\
F3        & Coolant inlet temperature           \\
F4        & Reactant inlet flow rate            \\
F5        & Reactant inlet concentration sensor \\
F6        & Reactant inlet temperature sensor   \\
F7        & Reactant outlet temperature sensor  \\
F8        & Coolant inlet flow rate sensor      \\
F9        & Coolant inlet temperature sensor    \\
F10       & Coolant outlet temperature sensor   \\
F11       & Catalyst decay                      \\
F12       & Heat transfer fouling               \\ \hline
\end{tabular}
\end{table}

\begin{table}[!ht]
\centering
\caption{Tasks setting of CSTR used in single source domain generalization fault diagnosis.}
\label{Table6}
\begin{tabular}{llll}
\hline
Task No. & \begin{tabular}[c]{@{}l@{}}Training set \\ (No. of samples:10400)\end{tabular} & \begin{tabular}[c]{@{}l@{}}Test set 1 \\ (No. of samples:2613 )\end{tabular} & \begin{tabular}[c]{@{}l@{}}Test set 2 \\ (No. of samples:26026 )\end{tabular} \\ \hline
T1       & M1                                                                             & M1                                                                           & M2,M3                                                                         \\
T2       & M2                                                                             & M2                                                                           & M1,M3                                                                         \\
T3       & M3                                                                             & M3                                                                           & M1,M2                                                                         \\ \hline
\end{tabular}
\end{table}

\subsection{Comparative models}
To validate the effectiveness of the DACN model, deep convolutional neural network (DCNN), CNN-LSTM, industrial process optimization ViT (IPO-ViT) and  multi-scale style generative and adversarial contrastive network(MSG-ACN) are used for comparative tests.

\begin{itemize}
    \item DCNN \cite{RN194629}. The DCNN model extracts the features in both spatial and temporal domains, consisting of convolutional layers, pooling layers, dropout, and fully connected layers.

\item  CNN-LSTM \cite{RN237022}. The CNN-LSTM model synthetically considers feature extraction and time delay of occurrence of faults. Feature extraction is achieved through CNN, and temporal delay is captured by LSTM.

\item  IPO-ViT\cite{RN237015}. The IPO-ViT model is built based on ViT, which utilizes the global receptive field provided by self-attention mechanism. This method can perform global feature learning on the complex process signals.

\item  MSG-ACN\cite{RN186301}. This method establishes a multi-scale style generation strategy to generate diverse samples. Then  domain-invariant features are extracted from the source and extended domains via an  adversarial contrastive learning strategy. It has been validated a relatively good performance in solving single-source domain generalization problems. 

\item  MSG-ACN2. The multi-scale style generation strategy and adversarial contrastive learning strategy in MSG-ACN are applied to pre-trained model. The multi-scale style generation strategy is used to generate pseudo-sample features instead of unseen mode samples there.

\end{itemize}

\subsection{Evaluation metrics}
The performance of the fault diagnosis model for class $l$ is measured using accuracy (ACC), fault diagnosis rate (FDR) and false positive rate (FPR), defined as follows,

\begin{equation}
    ACC=\frac{\sum_{l=1}^LTP_l}{\sum_{l=1}^LTP_l+\sum_{l=1}^LFN_l}
    \label{Eq9}
\end{equation}

\begin{equation}
    FDR_l=\frac{TP_l}{TP_l+FN_l}
    \label{Eq10}
\end{equation}

\begin{equation}
    FPR_l=\frac{FP_l}{FP_l+TN_l}
    \label{Eq11}
\end{equation}
where $TP_l$, $FN_l$, $FP_l$ and $TN_l$ are define in the confusion matrix for the $l$-th class, as shown in \hyperref[Table7]{Table 7}. Due to the inherent randomness in the training process, the accuracy of the model does fluctuate to some extent. In real-world applications, especially in the field of fault diagnosis, the reliability of the model is critical. Therefore, we conducted five independent training runs for each model and selected the lowest accuracy as the final result to ensure the model's performance is robust even in the worst-case scenario.

\begin{table}[!ht]
\centering
\caption{Confusion matrix for the $l$-th class(Number of samples).}
\label{Table7}

\begin{tabular}{lll}
\hline
                            & Predicted system health condition class is $l$ & Predicted system health condition class is not $l$ \\ \hline
Actual system health condition class is $l$     & $TP_l$                         & $FN_l$                             \\
Actual system health condition class is not $l$ & $FP_l$                         & $TN_l$                             \\ \hline
\end{tabular}
\end{table}

In addition to the above evaluation metrics, the model's parameter number, training time and testing time are also used to assess model performance.

\section{Experimental Results and Analysis}
\label{Sec5}
\subsection{TE process}
\subsubsection{Model comparison}
\begin{figure}[!ht]
    \centering
    \includegraphics[width=1.\textwidth,height=0.7212779\textwidth]{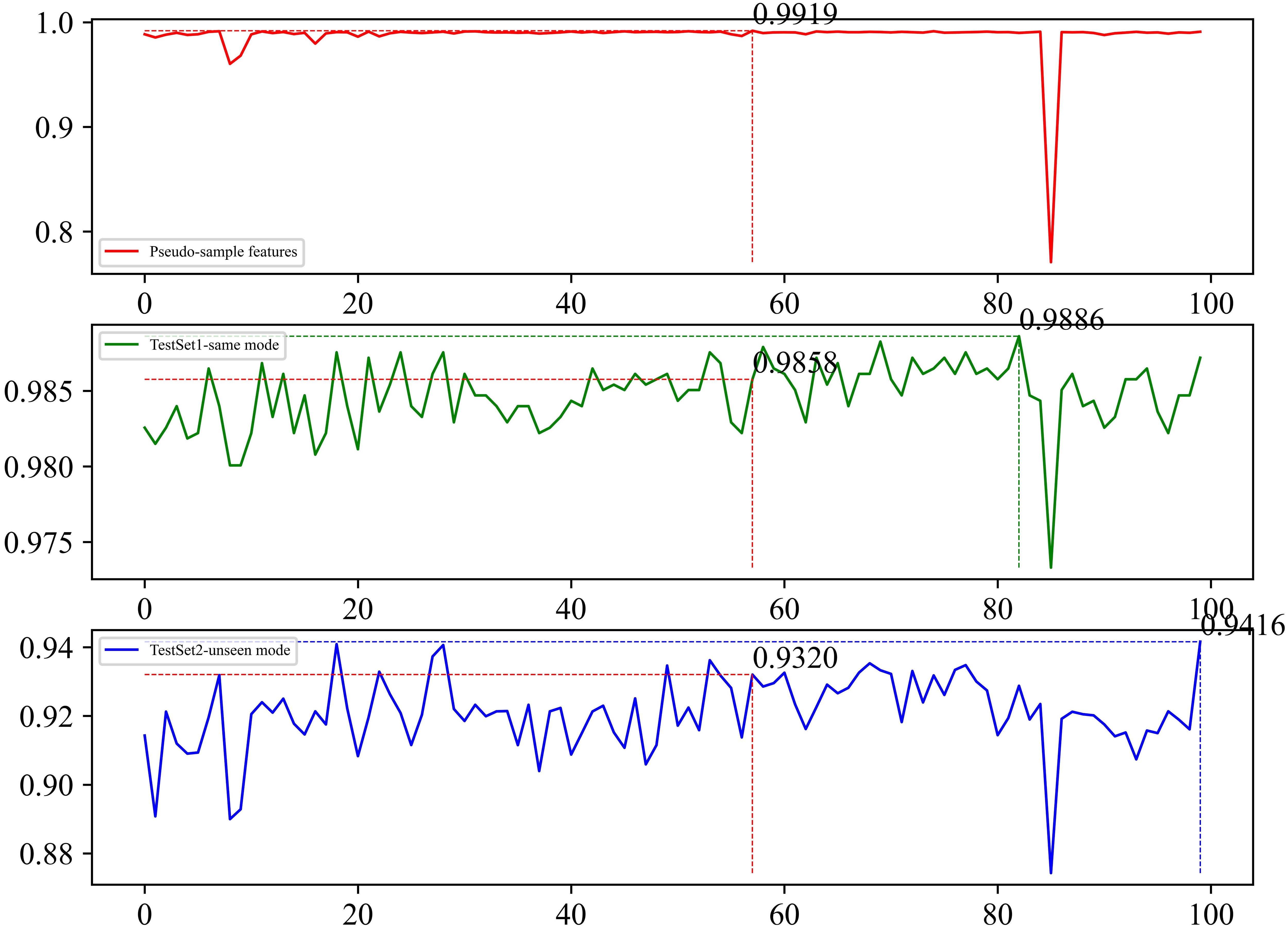}
    \caption{Classification accuracy on test set 1, test set 2, and pseudo-sample features during the Bayesian optimization process.}
    \label{Fig7}
\end{figure}

DCNN, CNN-LSTM, IPO-ViT, MSG-ACN, MSG-ACN2 and DACN were tested in the tasks shown in \hyperref[Table3]{Table 3} respectively. Among the methods for hyperparameter optimization, Bayesian optimization updates the prior iteratively during the search process, enabling faster convergence to the optimal hyperparameters. In contrast, other methods like grid search or random search are generally less efficient and slower in exploring the search space. Therefore, Bayesian optimization was chosen to determine the hyperparameters of MSG-ACN, MSG-ACN2, and DACN. For MSG-ACN, the optimization objective is to maximize the classification accuracy of generated samples, while for other models, the objective is to maximize the classification accuracy of pseudo-sample features. Taking T1 as an example, the classification accuracy of DACN on pseudo-sample features, test set 1 and test set 2 during the Bayesian optimization process are shown in \hyperref[Fig7]{Fig. 7}. It can be observed that the classification accuracy of pseudo-sample features reached its highest value of 99.19\% in the 57-th optimization iteration. At this point, the classification accuracy on the test set 1 is 98.58\%, and on test set 2 it is 93.20\%. During the optimization process, the maximum classification accuracies on the two test sets were 98.86\% and 94.16\%, respectively. This indicates that DACN has higher diagnostic potential in both the seen mode  and unseen modes.

To evaluate whether the model has basic diagnostic capabilities on data distributions similar to the training data, each model is tested on test set 1.
The accuracies achieved by each model under different tasks are shown in \hyperref[Table8]{Table 8}. CNN-LSTM obtained the highest average classification accuracy. However, the classification accuracies of CNN-LSTM, MSG-ACN and DACN are all higher than 99\%. This indicates that these three models exhibit high performance in fault diagnosis on single seen mode data. 

\begin{table}[!ht]
\centering
\caption{Experimental results on the test set 1 of TE process (Accuracy).}
\label{Table8}
\begin{tabular}{lllllll}
\hline
\textbf{Task No.} & \textbf{DCNN} & \textbf{CNN-LSTM} & \textbf{IPO-ViT} & \textbf{MSG-ACN} & \textbf{MSG-ACN2} & \textbf{DACN} \\ \hline
\textbf{T1}     & 97.54\%       & 99.07\%           & 97.94\%          & 99.00\%          & 98.22\%           & 98.68\%       \\
\textbf{T2}     & 88.22\%       & 99.40\%           & 97.12\%          & 99.40\%          & 98.83\%           & 99.25\%       \\
\textbf{T3}     & 97.76\%       & 99.50\%           & 98.36\%          & 99.36\%          & 98.33\%           & 99.32\%       \\
\textbf{T4}     & 87.58\%       & 98.97\%           & 96.83\%          & 98.68\%          & 97.37\%           & 98.54\%       \\
\textbf{T5}     & 98.04\%       & 99.40\%           & 97.26\%          & 99.32\%          & 99.07\%           & 99.29\%       \\
\textbf{T6}     & 96.90\%       & 99.32\%           & 98.01\%          & 99.18\%          & 98.43\%           & 99.07\%       \\\hline
\textbf{Avg}    & 94.34\%       & \textbf{99.28\%}  & 97.59\%          & 99.16\%          & 98.38\%           & 99.03\%       \\ \hline
\end{tabular}
\end{table}

The model's generalization performance on unseen modes is then evaluated. For the test set 2 of TE process, the accuracies achieved by each model under different tasks are shown in \hyperref[Table9]{Table 9}. DACN attains the highest classification accuracy in 5 out of 6 tasks and obtained the highest average classification accuracy. The comparison between MSG-ACN and MSG-ACN2 indicates that in the TE process scenario, generating pseudo-sample features instead of unseen mode samples does not yield better performance. However, the proposed model DACN still achieves better average classification accuracy, confirming the effectiveness of supervised contrastive learning and adversarial learning in promoting domain-invariant feature extraction. Although CNN-LSTM, MSG-ACN, and DACN all achieve an average classification accuracy of over 99\% on single seen mode data, DACN's average classification accuracy on unseen modes is 2.99\% and 1.3\% higher than that of CNN-LSTM and MSG-ACN, respectively. This indicates that DACN has superior performance in extracting domain-invariant features. MSG-ACN has been validated to possess superior fault diagnosis performance on unseen modes \cite{RN186301}. DACN's average classification accuracy on test set 2 is still higher than that of MSG-ACN, demonstrating DACN's excellent diagnostic performance on unseen modes.

\begin{table}[!ht]
\centering
\caption{Experimental results on the test set 2 of TE process (Accuracy).}
\label{Table9}

\begin{tabular}{lllllll}
\hline
\textbf{Task No.} & \textbf{DCNN} & \textbf{CNN-LSTM} & \textbf{IPO-ViT} & \textbf{MSG-ACN} & \textbf{MSG-ACN2} & \textbf{DACN}    \\ \hline
\textbf{T1}     & 87.24\%       & 90.12\%           & 86.02\%          & 92.37\%          & 89.50\%           & \textbf{92.86\%} \\
\textbf{T2}     & 76.89\%       & 88.75\%           & 88.66\%          & 91.06\%          & 90.61\%           & \textbf{93.01\%} \\
\textbf{T3}     & 81.92\%       & 83.84\%           & 75.41\%          & 85.49\%          & 82.48\%           & \textbf{85.57\%} \\
\textbf{T4}     & 75.32\%       & 87.85\%           & 88.88\%          & 87.52\%          & 87.34\%           & \textbf{91.66\%} \\
\textbf{T5}     & 88.83\%       & 88.35\%           & 88.97\%          & 91.31\%          & 91.54\%           & \textbf{93.14\%} \\
\textbf{T6}     & 84.39\%       & 85.79\%           & 78.63\%          & \textbf{87.09\%} & 84.71\%           & 86.38\%           \\\hline
\textbf{Avg}    & 82.43\%       & 87.45\%           & 84.43\%          & 89.14\%          & 87.70\%           & \textbf{90.44\%} \\ \hline
\end{tabular}
\end{table}

\begin{figure}[!ht]
    \centering
    \includegraphics[width=0.6\textwidth,height=0.4025\textwidth]{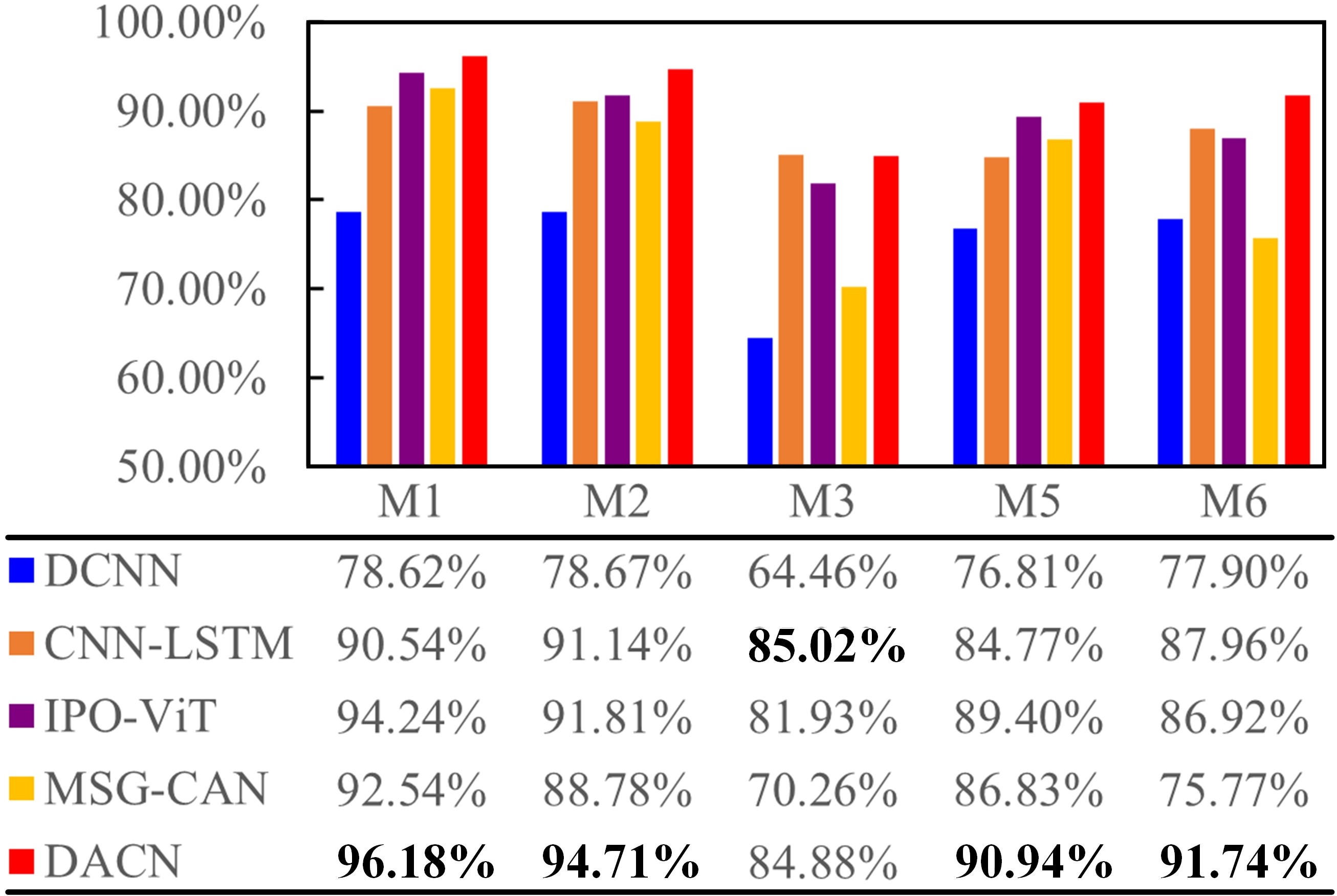}
    \caption{ACC of the trained models for each mode (M1, M2, M3, M5 and M6) in task T4.}
    \label{Fig8}
\end{figure}

Taking task T4 as an example, the FDR and FPR obtained by each model are shown in \hyperref[Table10]{Table 10}. DACN attains the highest average FDR and the lowest average FPR. Additionally, the diagnostic accuracies of the trained models for each mode (M1, M2, M3, M5 and M6) is shown in \hyperref[Fig8]{Fig. 8}. DACN achieves the highest classification accuracy in four unseen modes. Although its classification accuracy in M3 is slightly lower than that of CNN-LSTM, the difference is small. DACN exhibits outstanding overall performance, which proves its ability to effectively generalize to unseen modes. 

\begin{table}[!ht]
\centering
\caption{FDR and FPR of DCNN, CNN-LSTM, IPO-ViT, MSG-ACN and DACN in task T4.}
\label{Table10}
\small
\begin{tabular}{lllllllllll}
\hline
      & \multicolumn{5}{l}{FDR}  & \multicolumn{5}{l}{FPR}                                                                                                                                                                  \\ \hline
\begin{tabular}[c]{@{}l@{}}Fault \\ No.\end{tabular} & DCNN   & \begin{tabular}[c]{@{}l@{}}CNN-\\ LSTM\end{tabular} & \begin{tabular}[c]{@{}l@{}}IPO-\\ ViT\end{tabular} & \begin{tabular}[c]{@{}l@{}}MSG-\\ ACN\end{tabular} & DACN            & DCNN   & \begin{tabular}[c]{@{}l@{}}CNN-\\ LSTM\end{tabular} & \begin{tabular}[c]{@{}l@{}}IPO-\\ ViT\end{tabular} & \begin{tabular}[c]{@{}l@{}}MSG-\\ ACN\end{tabular} & DACN            \\ \hline
F0                                                   & 0.9732 & 0.9620                                              & 0.8858                                             & 0.7675                                             & 0.9586          & 0.0683 & 0.0353                                              & 0.0359                                             & 0.0140                                             & 0.0422          \\
F2                                                   & 0.9223 & 0.8728                                              & 0.9762                                             & 0.8448                                             & 0.9071          & 0.0000 & 0.0013                                              & 0.0054                                             & 0.0007                                             & 0.0016          \\
F4                                                   & 0.9753 & 0.9979                                              & 0.9953                                             & 0.9966                                             & 0.9957          & 0.0000 & 0.0014                                              & 0.0002                                             & 0.0041                                             & 0.0001          \\
F8                                                   & 0.7238 & 0.8310                                              & 0.6734                                             & 0.7829                                             & 0.8288          & 0.0047 & 0.0216                                              & 0.0092                                             & 0.0129                                             & 0.0045          \\
F10                                                  & 0.9723 & 0.9759                                              & 0.8915                                             & 0.9787                                             & 0.9763          & 0.0337 & 0.0285                                              & 0.0228                                             & 0.0431                                             & 0.0273          \\
F11                                                  & 0.9677 & 0.9839                                              & 0.9877                                             & 0.9887                                             & 0.9926          & 0.0014 & 0.0010                                              & 0.0007                                             & 0.0015                                             & 0.0012          \\
F12                                                  & 0.3359 & 0.3680                                              & 0.7475                                             & 0.7589                                             & 0.7833          & 0.0163 & 0.0075                                              & 0.0013                                             & 0.0130                                             & 0.0055          \\
F13                                                  & 0.6791 & 0.8164                                              & 0.7535                                             & 0.9122                                             & 0.8398          & 0.0071 & 0.0347                                              & 0.0473                                             & 0.0234                                             & 0.0103          \\
F14                                                  & 0.0000 & 0.9892                                              & 0.9892                                             & 0.7342                                             & 0.8979          & 0.0000 & 0.0028                                              & 0.0007                                             & 0.0062                                             & 0.0001          \\
F17                                                  & 0.9823 & 0.9876                                              & 0.9877                                             & 0.9873                                             & 0.9857          & 0.1426 & 0.0008                                              & 0.0000                                             & 0.0199                                             & 0.0000          \\\hline
Avg                                                  & 0.7532 & 0.8785                                              & 0.8888                                             & 0.8752                                             & \textbf{0.9166} & 0.0274 & 0.0135                                              & 0.0124                                             & 0.0139                                             & \textbf{0.0093} \\ \hline
\end{tabular}
\end{table}

T-SNE is used to visualize the feature maps. Taking task T1 as an example, M2 from test set 2 is selected for visualization. 

\begin{figure}[!ht]
    \centering
    \includegraphics[width=1.\textwidth,height=1.1598\textwidth]{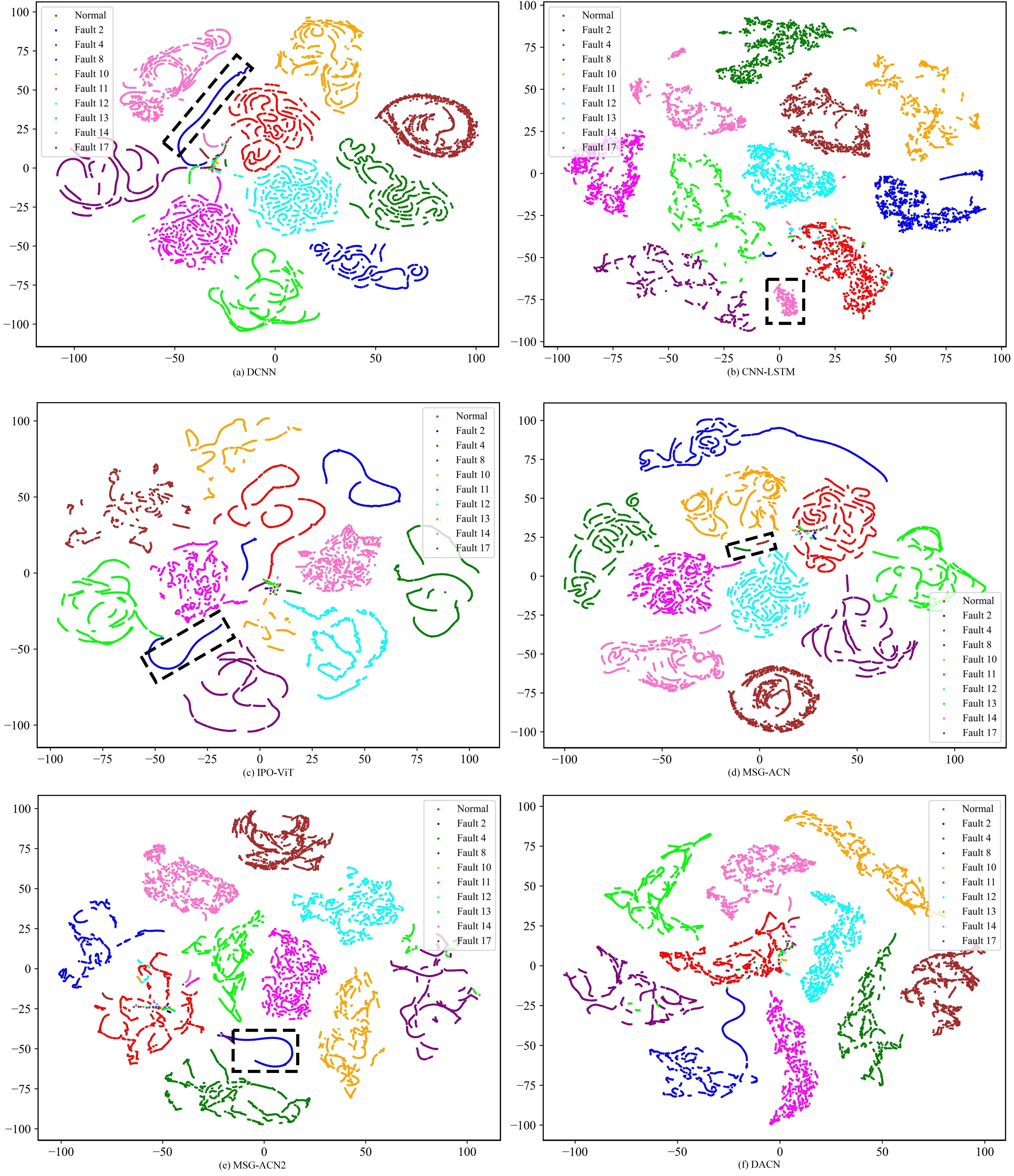}
    \caption{Feature visualization using T-SNE in task T1.}
    \label{Fig9}
\end{figure}

The results for each model are shown in \hyperref[Fig9]{Fig. 9}. Colors represent system health condition classes, with a total of ten. The black dashed box highlights a set of samples that are clearly misclassified into other clusters. Such as Fault 2 samples in \hyperref[Fig9]{Fig. 9} (a), (c) and (e), Fault 4 and Fault 17 samples in \hyperref[Fig9]{Fig. 9} (d), and Fault 14 samples in \hyperref[Fig9]{Fig. 9} (b). Compared to other methods, the features extracted by DACN effectively separate the boundaries between different classes, with no significant overlap between different clusters. This facilitates the differentiation of system health condition classes, thereby validating the high generalization performance of DACN.

\hyperref[Table11]{Table 11} presents the parameter number, training and inference times of DCNN, CNN-LSTM, IPO ViT, MSG-ACN, MSG-ACN2 and DACN models. Although DACN requires more parameters during training process compared to CNN-LSTM and MSG-ACN, it achieves shorter training time. {In the CNN-LSTM model, the sequence dependency within the LSTM limits parallelized processing, leading to an increase in training time. In MSG-CAN and MSG-ACN2 models, the parameter update process is divided into two steps as detailed in \cite{RN186301}, , which increases the training time. In contrast, the DACN model avoids these limitations as it does not involve sequential dependencies or multi-stage parameter updates. This makes the training of the DACN model more efficient.} In the inference process, both DACN and MSG-ACN demonstrate the shortest time consumption. Overall, compared to other models, DACN demonstrates higher computational efficiency, validating its advantage in practical deployment.

\begin{table}[!ht]
\centering
\caption{Parameter number, training and inference time spent of each model.}
\label{Table11}
\small
\begin{tabular}{lllllll}
\hline
Evaluation metrics     & DCNN       & \begin{tabular}[c]{@{}l@{}}CNN-\\ LSTM\end{tabular} & \begin{tabular}[c]{@{}l@{}}IPO-\\ ViT\end{tabular} & \begin{tabular}[c]{@{}l@{}}MSG-\\ ACN\end{tabular}   & \begin{tabular}[c]{@{}l@{}}MSG-\\ ACN2\end{tabular} & DACN      \\ \hline
\begin{tabular}[c]{@{}l@{}}No. of parameters \\ in training process\end{tabular}  & 11,404,334 & 895,330                                             & 26,378,350                                         & 680,034 & 2,316,426                                          & 1,944,039 \\
\begin{tabular}[c]{@{}l@{}}Time spent on model\\  training (s/epoch)\end{tabular} & 3.76       & 3.07                                                & 10.33                                              & 3.06    & 3.08                                               & 2.63      \\
\begin{tabular}[c]{@{}l@{}}No. of parameters \\ in inference process\end{tabular} & 11,404,334 & 895,330                                             & 26,378,350                                         & 645,002 & 645,002                                            & 645,002   \\
\begin{tabular}[c]{@{}l@{}}Time spent on \\ inference (s/epoch)\end{tabular}      & 1.09       & 0.06                                                & 8.35                                               & 0.01    & 0.01                                               & 0.01      \\ \hline
\end{tabular}
\end{table}
\subsubsection{Ablation study}
Ablation experiments were conducted to verify the effectiveness of pre-training, supervised contrastive learning, and adversarial learning. The pre-trained model is denoted as A1. A2 and A3 represent models where the supervised contrastive loss and discriminative loss are removed from DACN, respectively. A4 represents DACN without the pre-training process, optimized directly through training process. The hyperparameters for these models were obtained using Bayesian optimization, targeting the highest classification accuracy for pseudo-sample features. The classification accuracies of these different models on test set 2 are shown in \hyperref[Table12]{Table 12}.

\begin{table}[!ht]
\centering
\caption{Results of ablation experiments.}
\label{Table12}
\small
\begin{tabular}{llllll}
\hline
\textbf{Task No.} & \textbf{A1} & \textbf{A2} & \textbf{A3} & \textbf{A4}      & \textbf{DACN}    \\ \hline
\textbf{T1}       & 85.75\%     & 91.76\%     & 92.17\%     & 91.95\%          & \textbf{92.86\%} \\
\textbf{T2}       & 86.52\%     & 92.44\%     & 91.84\%     & 92.61\%          & \textbf{93.01\%} \\
\textbf{T3}       & 79.46\%     & 84.19\%     & 82.19\%     & \textbf{86.72\%} & 85.57\%          \\
\textbf{T4}       & 80.30\%     & 90.53\%     & 91.66\%     & 91.53\%          & \textbf{91.66\%} \\
\textbf{T5}       & 86.87\%     & 91.23\%     & 89.88\%     & 91.56\%          & \textbf{93.14\%} \\
\textbf{T6}       & 83.22\%     & 86.12\%     & 86.38\%     & 85.99\%          & \textbf{86.38\%} \\\hline
\textbf{Avg}      & 83.69\%     & 89.38\%     & 89.02\%     & 90.06\%          & \textbf{90.44\%} \\ \hline
\end{tabular}
\end{table}

A2, A3 and DACN are all improvements based on A1. They all show increased average classification accuracy compared to A1.  This indicates that both discriminator and supervised contrastive learning contribute to the model's generalization ability, and their combination achieved the optimal diagnostic results. DACN has an average classification accuracy 0.38\% higher than A4. This indicates that the pre-trained model can effectively guide to generate valid pseudo-sample features. In TE process scenario, pre-training, supervised
contrastive learning, and adversarial learning strategies all contribute to enhancing the model's ability to extract domain-invariant features. This confirms the high applicability of DACN for fault diagnosis on unseen modes.

\subsection{CSTR}
\subsubsection{Model comparison}

\begin{figure}[!ht]
    \centering
    \includegraphics[width=1.\textwidth,height=0.7464\textwidth]{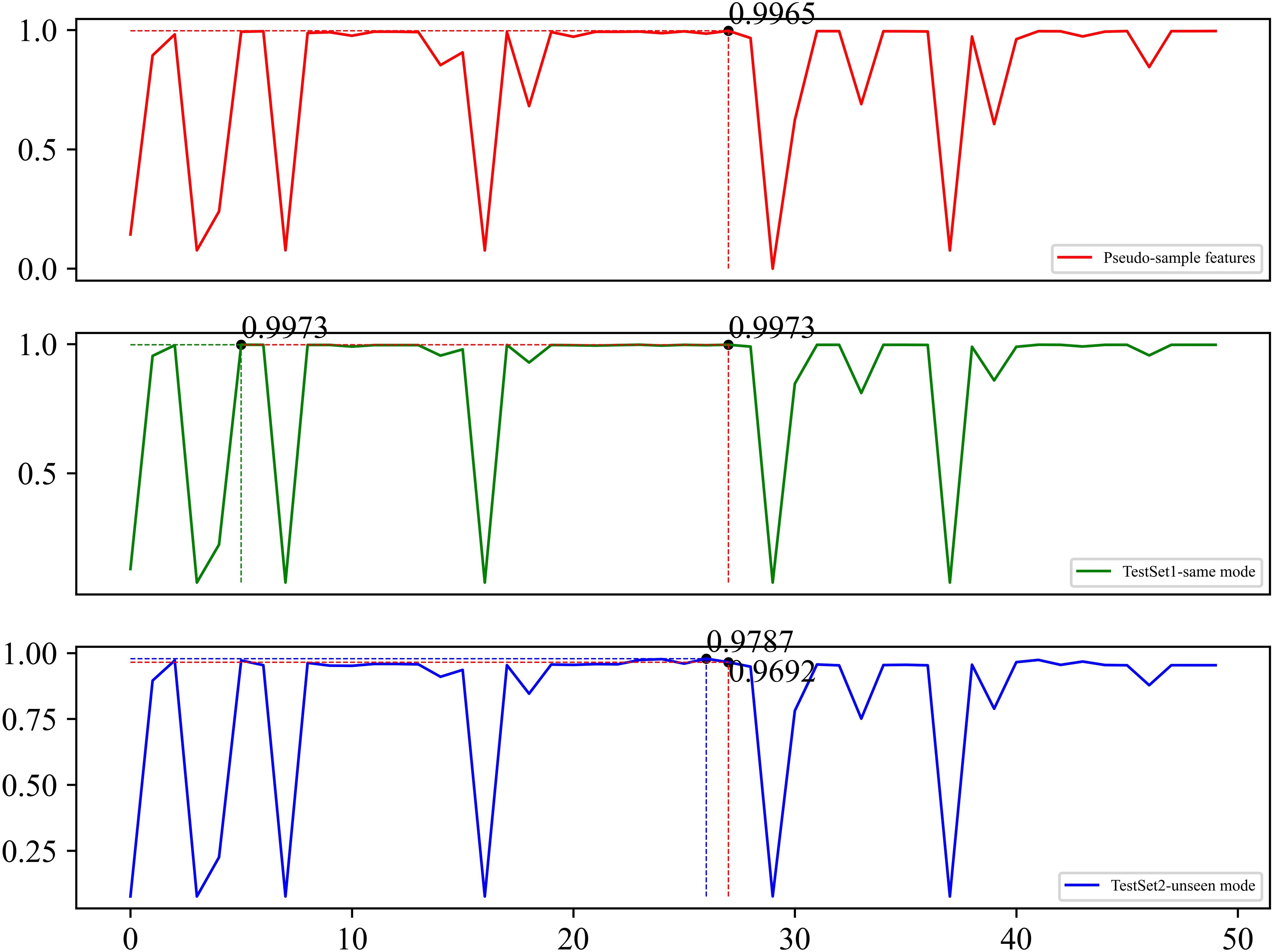}
    \caption{Classification accuracy on test set 1, test set 2, and pseudo-sample features during the Bayesian optimization process. ( Note that during the iteration process, some classification accuracies are less than 50\%. This is due to the loss value becoming NaN, causing the optimization process to be interrupted.)}
    \label{Fig10}
\end{figure}

Similar to TE experiment, six models were tested in the tasks shown in \hyperref[Table5]{Table 5} respectively. Taking T3 as an example, the classification accuracy of DACN on pseudo-sample features, test set 1 and test set 2 during the Bayesian optimization process are shown in \hyperref[Fig10]{Fig. 10}. It can be observed that the maximum classification accuracy of pseudo-sample features is 99.65\%. At this point, the classification accuracy on the test set 1 is 99.73\%, and on test set 2 it is 96.92\%. However, during the optimization process, the maximum classification accuracy on test set 2 is 97.87\%. The results indicate that the optimized model still exhibits single-domain overfitting, and its generalization ability to unseen modes can be further improved. 

For the test set 1 of CSTR, the accuracies achieved by each model under different tasks
are shown in \hyperref[Table13]{Table 13}. DACN obtained the highest average classification accuracy. CNN-LSTM, MSG-ACN, MSG-ACN2 and DACN all attained average classification accuracies higher than 99\%, indicating that they have successfully acquired knowledge of single-mode faults in CSTR. \hyperref[Table14]{Table 14} shows the experimental results on the test set 2. The average classification accuracy of MSG-ACN2 is higher than that of MSG-ACN, indicating that generating pseudo-sample features is more effective than generating unseen mode samples in CSTR scenario. MSG-ACN did not achieve good diagnostic performance in the CSTR scenario. This is hypothesized to be due to the small differences between some faulty and normal samples, making it difficult for the model to capture deep domain-invariant features. MSG-ACN2 achieved the highest classification accuracy in task T1. However, its average classification accuracy was 0.91\% lower than that of DACN. This indicates that MSG-ACN2 can only enhance the model's generalization ability under specific unseen modes. DACN
achieved the highest classification accuracy in all three tasks, as well as the highest average classification accuracy. The comprehensive view shows that DACN has superior diagnostic performance on unseen modes, confirming its outstanding ability to extract domain-invariant features.

\begin{table}[!ht]
\centering
\caption{Experimental results on the test set 1 of CSTR (Accuracy).}
\label{Table13}
\small
\begin{tabular}{lllllll}
\hline
\textbf{Task No.} & \textbf{DCNN} & \textbf{CNN-LSTM} & \textbf{IPO-ViT} & \textbf{MSG-ACN} & \textbf{MSG-ACN2} & \textbf{DACN}    \\ \hline
\textbf{T1}       & 89.82\%       & 99.39\%           & 98.32\%          & 99.31\%          & 99.39\%           & 99.69\%          \\
\textbf{T2}       & 88.27\%       & 99.35\%           & 98.97\%          & 99.31\%          & 99.46\%           & 99.77\%          \\
\textbf{T3}       & 91.20\%       & 99.31\%           & 98.09\%          & 99.69\%          & 99.50\%           & 99.69\%          \\\hline
\textbf{Avg}      & 89.76\%       & 99.35\%           & 98.46\%          & 99.44\%          & 99.45\%           & \textbf{99.72\%} \\ \hline
\end{tabular}
\end{table}

\begin{table}[!ht]
\centering
\caption{Experimental results on the test set 2 of CSTR (Accuracy).}
\label{Table14}
\small
\begin{tabular}{lllllll}
\hline
\textbf{Task No.} & \textbf{DCNN} & \textbf{CNN-LSTM} & \textbf{IPO-ViT} & \textbf{MSG-ACN} & \textbf{MSG-ACN2} & \textbf{DACN}    \\ \hline
\textbf{T1}       & 87.18\%       & 92.23\%           & 95.49\%          & 95.07\%          & \textbf{97.66\%}  & \textbf{97.66\%}          \\
\textbf{T2}       & 87.58\%       & 95.19\%           & 95.91\%          & 96.86\%          & 97.46\%           & \textbf{98.20\%} \\
\textbf{T3}       & 85.55\%       & 93.75\%           & 93.16\%          & 90.02\%          & 94.91\%           & \textbf{96.92\%} \\\hline
\textbf{Avg}      & 86.77\%       & 93.72\%           & 94.85\%          & 93.98\%          & 96.68\%           & \textbf{97.59\%} \\ \hline
\end{tabular}
\end{table}

\begin{figure}[!ht]
    \centering
    \includegraphics[width=0.8\textwidth,height=1.13984\textwidth]{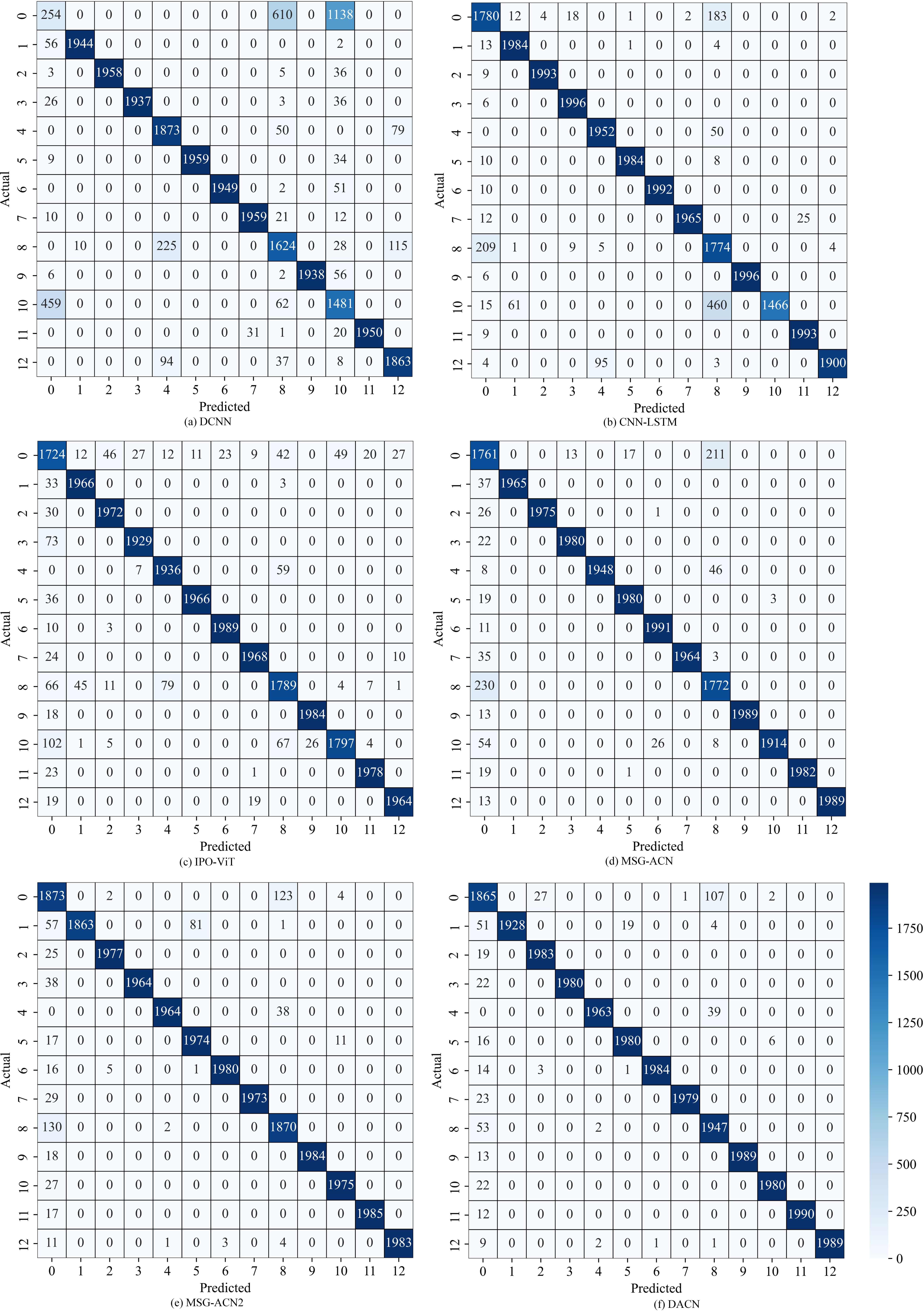}
    \caption{Confusion matrix for each model in the T2 task.}
    \label{Fig11}
\end{figure}

\begin{figure}[!ht]
    \centering
    \includegraphics[width=1.\textwidth,height=1.1556\textwidth]{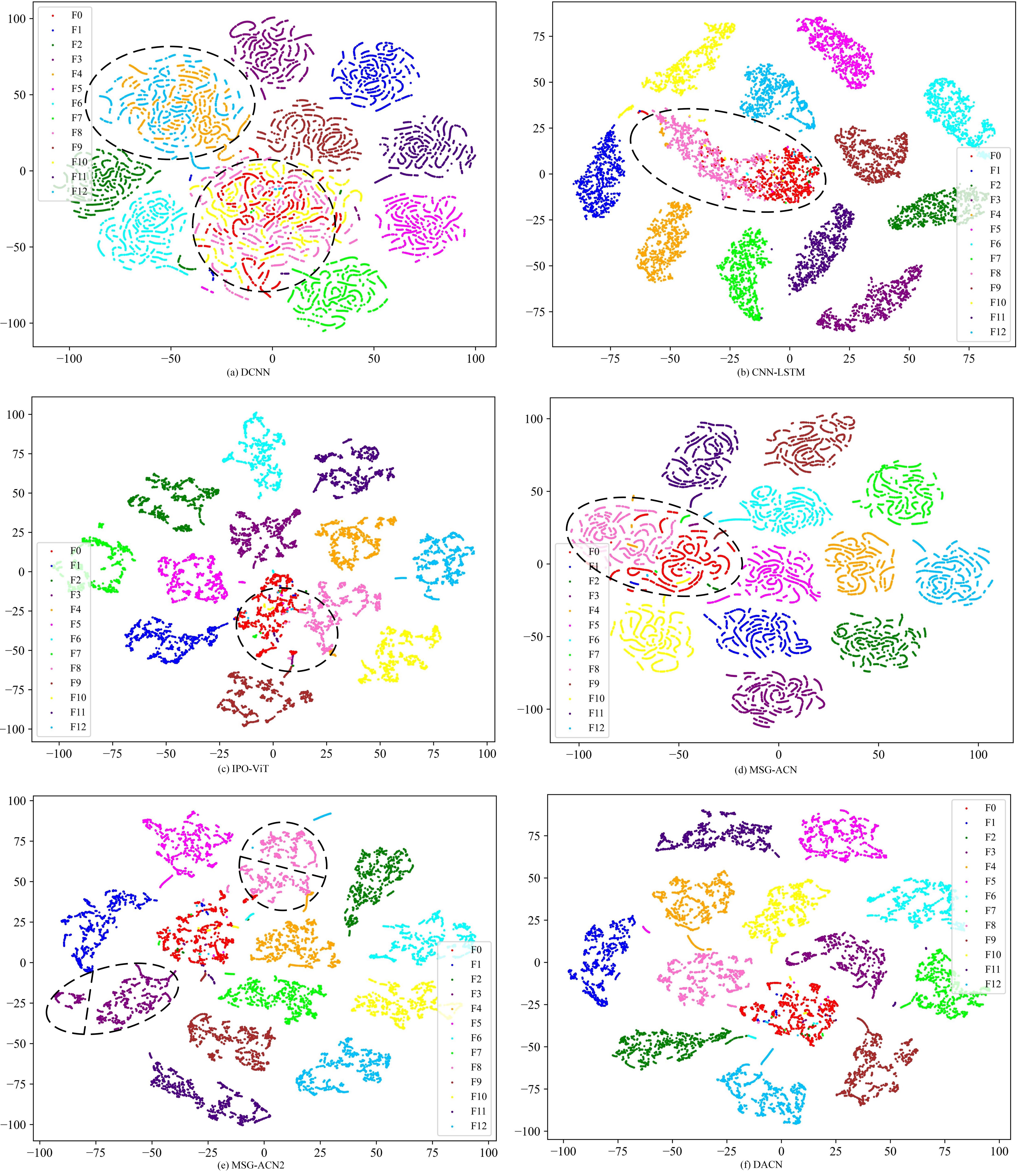}
    \caption{Feature visualization using T-SNE in task T2.}
    \label{Fig12}
\end{figure}

Taking task T2 as an example, the confusion matrix for each model is shown in \hyperref[Fig11]{Fig. 11}. The number of samples correctly classified by the DACN is large, demonstrating the excellent performance of DACN. M1 from test set 2 is selected for feature map visualization, as shown in \hyperref[Fig12]{Fig. 12}. Thirteen colors represent thirteen different system health condition classes. In \hyperref[Fig12]{Fig. 12} (a), (b), (c), and (d), the dashed lines encircle a large number of overlapping samples from different classes. Such as F4 and F12 in \hyperref[Fig12]{Fig. 12} (a), F8 and F0 in \hyperref[Fig12]{Fig. 12} (b), (c) and (d). In \hyperref[Fig12]{Fig. 12}(e), the dashed lines encircle the samples with low aggregation, which can be separated by the internal dashed line. As shown in \hyperref[Fig12]{Fig. 12}(f),  DACN effectively distinguishes system health condition classes, with each class being more concentrated. Therefore, the proposed DACN demonstrates good performance. 

The parameter number, training and inference times of each model are counted, as shown in \hyperref[Table15]{Table 15}. {The experimental results are similar to those obtained in TE process. The comprehensive comparison of the training and testing times across different models further validates the competitive advantage of DACN in practical applications.}

\begin{table}[!ht]
\centering
\caption{Parameter number, training and inference time spent of each model.}
\label{Table15}
\small
\begin{tabular}{llllllll}
\hline
Evaluation metrics                                                                            & DCNN      & \begin{tabular}[c]{@{}l@{}}CNN-\\ LSTM\end{tabular} & \begin{tabular}[c]{@{}l@{}}IPO-\\ ViT\end{tabular} & \begin{tabular}[c]{@{}l@{}}MSG-\\ ACN\end{tabular} & \begin{tabular}[c]{@{}l@{}}MSG-\\ ACN2\end{tabular} & DACN      \\ \hline
\begin{tabular}[c]{@{}l@{}}No. of parameters\\ in training process\end{tabular}  & 2,266,037 & 578,333                                             & 25,290,785                                          & 649,221                                            & 692,109                                             & 2,161,386 \\
\begin{tabular}[c]{@{}l@{}}Time spent on model\\ training (s/epoch)\end{tabular} & 2.49      & 3.04                                                & 9.89                                                   & 2.86                                               & 2.83                                                & 2.53      \\
\begin{tabular}[c]{@{}l@{}}No. of parameters in\\ inference process\end{tabular} & 2,266,037 & 578,333                                             & 25,290,785                                         & 630,413 & 630,413                  & 630,413   \\
\begin{tabular}[c]{@{}l@{}}Time spent on\\ inference (s/epoch)\end{tabular}      & 0.019     & 0.027                                               & 8.67                                                  & 0.006                                              & 0.006                                               & 0.006     \\ \hline
\end{tabular}
\end{table}

\subsubsection{Ablation study}
The models used for the ablation study in the CSTR are consistent with those used in the TE process. The experimental results of each model on the test set 2 are shown in \hyperref[Table16]{Table 16}. 

Comparing the classification accuracies of A1, A2, A3, and DACN, it is evident that in the CSTR scenario, supervised contrastive learning consistently enhances diagnostic performance for unseen modes. Adversarial learning only improves the model's generalization performance under certain specific mode. Combining both approaches yields the best diagnostic results. 
This indicates that supervised contrastive learning and adversarial learning have a mutually reinforcing effect, enhancing the model's ability to learn domain-invariant features.
The average classification accuracy of DACN is 1.55\% higher relative to A4, which confirms the function of pre-training. It helps to generate valid pseudo-sample features. Overall, combining pre-training strategies, supervised contrastive learning strategies, and adversarial learning strategies achieved the best average classification accuracy, verifying the model's excellent generalization ability on unseen modes.

\begin{table}[!ht]
\centering
\caption{Results of ablation experiments.}
\label{Table16}
\small
\begin{tabular}{llllll}
\hline
\textbf{Task No.} & \textbf{A1} & \textbf{A2} & \textbf{A3} & \textbf{A4}      & \textbf{DACN}    \\ \hline
\textbf{T1}       & 97.19\%     & 97.60\%     & 96.68\%     & 97.40\%          & \textbf{97.66\%} \\
\textbf{T2}       & 97.61\%     & 98.03\%     & 97.67\%     & \textbf{98.21\%} & 98.20\%          \\
\textbf{T3}       & 96.07\%     & 96.67\%     & 95.30\%     & 92.52\%          & \textbf{96.92\%} \\\hline
\textbf{Avg}      & 96.96\%     & 97.43\%     & 96.55\%     & 96.04\%          & \textbf{97.59\%} \\ \hline
\end{tabular}
\end{table}

\subsection{Hyperparameter and data volume sensitivity analysis}

\begin{figure}[!ht]
    \centering
    \includegraphics[width=1.\textwidth,height=0.7342\textwidth]{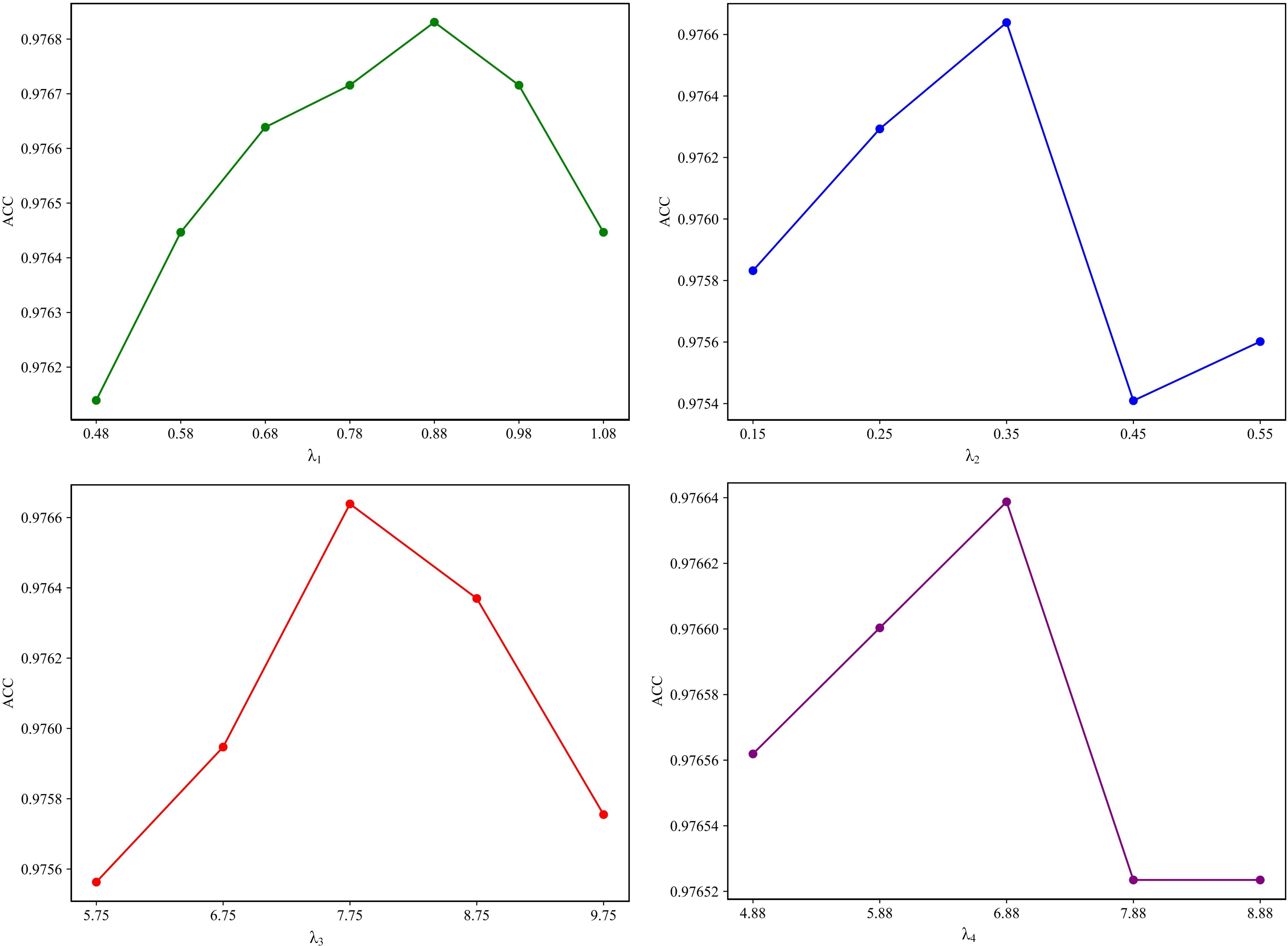}
    \caption{Experimental results under different hyperparameters.}
    \label{Fig13}
\end{figure}

To compare the impact of different hyperparameters on model performance, we conducted experiments using the T1 task of the CSTR dataset. The optimal hyperparameters $\lambda_1$, $\lambda_2$, $\lambda_3$ and $\lambda_4$ obtained through Bayesian optimization were 0.68, 0.35, 7.75, and 6.88, respectively. Further experiments were performed by adjusting the hyperparameters in the vicinity of these optimized values, and the classification accuracy of DACN on test set 2 is shown in \hyperref[Fig13]{Fig. 13}. It can be observed that as the hyperparameter values increase, the classification accuracy first increases and then decreases. Hyperparameters $\lambda_1$ and $\lambda_2$ represent the degree of attention model DACN pays to the single seen mode data and pseudo-sample features, respectively. Smaller values may prevent the model from capturing deep features in the data, while larger values can cause the model to overfit specific data or 'overshoot' the optimal classification boundary. Strong domain-invariant feature extraction capabilities and diverse pseudo-sample features, determined by $\lambda_3$ and $\lambda_4$, are crucial for training DACN. However, there is a trade-off between domain-invariant feature extraction and pseudo-sample feature diversity, requiring a balance between the two for optimal model performance. The classification accuracy remains relatively stable across different hyperparameter settings, indicating the model's robustness around the optimized values. Although Bayesian optimization yielded a $\lambda_1$ value of 0.68, experiments showed that when $\lambda_1$ is set to 0.88, the classification accuracy on test set 2 is the highest. This is because Bayesian optimization aimed to maximize the classification accuracy of pseudo-sample features, and while the optimized model did not achieve the highest accuracy on test set 2, it still outperformed other comparative models, further validating the superiority of DACN.

\begin{figure}[!ht]
    \centering
    \includegraphics[width=0.8\textwidth,height=0.59712\textwidth]{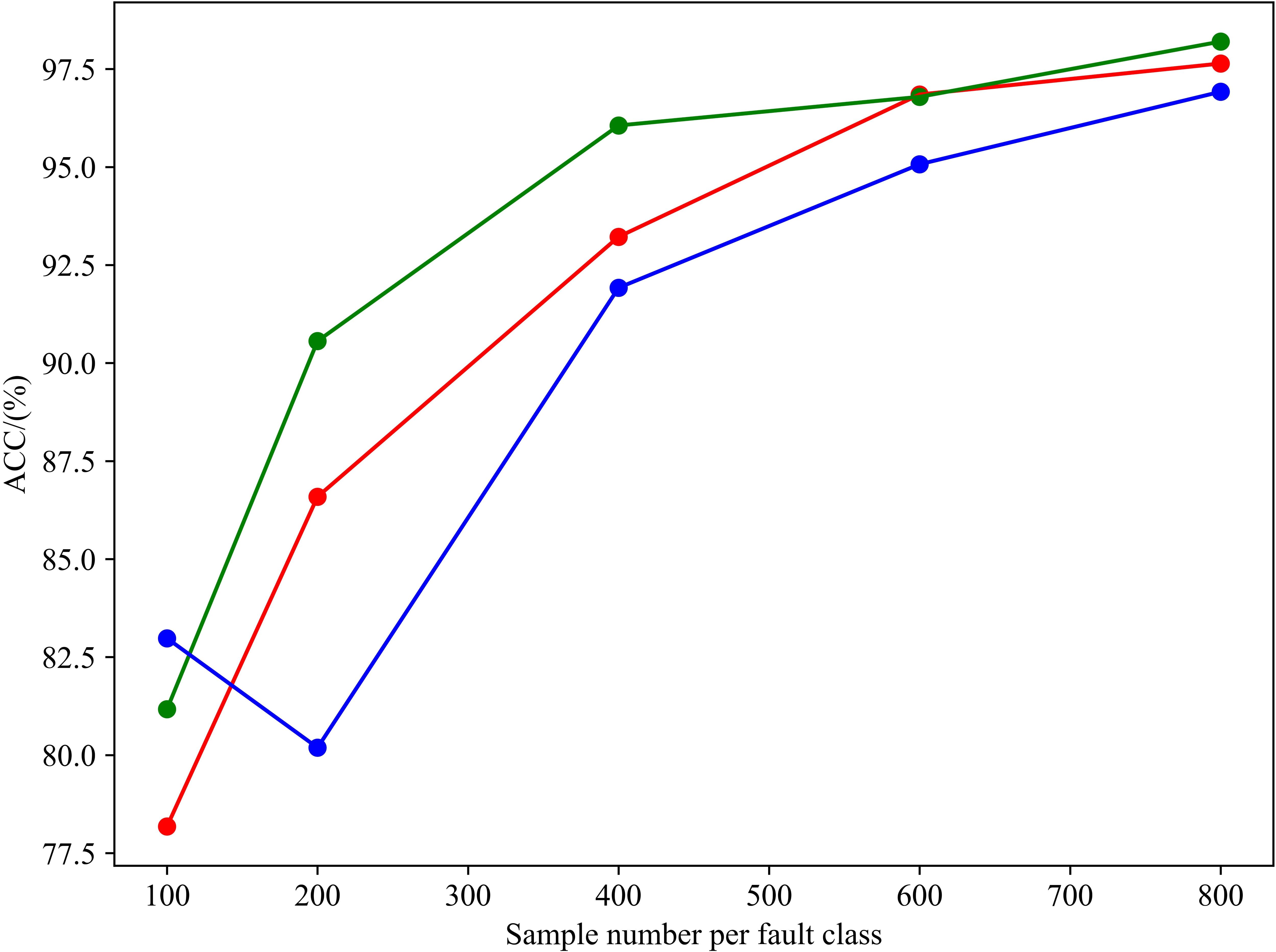}
    \caption{Experimental results under different data volumes.}
    \label{Fig14}
\end{figure}

To assess the impact of different data volumes on model performance, three tasks from the CSTR were used as examples, with the number of samples per system health condition class set to 200, 400, 600, and 800, respectively. The classification accuracy of DACN on test set 2 is presented in \hyperref[Fig14]{Fig. 14}. As the data volume increased, the classification accuracy of DACN also improved. When the number of training samples per system health condition class reached 400 or more, the accuracy of DACN exceeded 90\% across all tasks. However, with only 200 samples, the accuracy dropped below 90\% for task T2 and T3. These results indicate that DACN requires a sufficient number of samples to perform optimally and is less suitable for scenarios with limited data availability.

\section{Conclusion}
\label{Sec6}
In this paper, a single-source domain generalization model in fault diagnosis for process industrial systems is proposed. The novel method DACN can generate pseudo-sample features with sufficient semantic consistency and domain diversity, which represent sample features from unseen modes. Meanwhile, domain-invariant feature representation is learned from single seen mode sample features and pseudo-sample features, which can be applied to diagnose unseen mode samples. The experimental results on TE process and CSTR demonstrate the effectiveness of DACN. Some conclusions are drawn below.
\begin{itemize}

\item The adversarial pseudo-sample feature generation strategy produces fake unseen mode sample features with rich semantic information and diversity.  

\item Pre-training process can pre-extract sample features from a single mode. Then pseudo-sample features are generated by transforming the pre-extract sample features. This prevents the generation of useless pseudo-sample features.

\item The enhanced domain-invariant feature extraction strategy learns common feature representations of system health conditions, enabling robust fault diagnosis across different modes. 

\item DACN achieves robust generalization performance with fewer model parameters overall.

\end{itemize}

However, this paper considers the scenario where sample features from multiple modes share common domain-invariant features. 
Future work can address the issue of the partial differences in shared features between single seen mode and multiple unseen modes. The proposed model DACN still requires a sufficient amount of samples for training, making it unsuitable for applications with a small sample size.
Furthermore, the presence of unknown faults and class imbalance in unseen modes are issues that require further research.

\bibliographystyle{elsarticle-num} 
\bibliography{Main} 

\begin{thebibliography}{10}
\expandafter\ifx\csname url\endcsname\relax
  \def\url#1{\texttt{#1}}\fi
\expandafter\ifx\csname urlprefix\endcsname\relax\def\urlprefix{URL }\fi
\expandafter\ifx\csname href\endcsname\relax
  \def\href#1#2{#2} \def\path#1{#1}\fi

\bibitem{RN192229}
W.~J. Li, H.~Li, S.~Gu, T.~Chen, Process fault diagnosis with model- and
  knowledge-based approaches: Advances and opportunities, Control Engineering
  Practice 105 (2020) 17.
\newblock \href {https://doi.org/10.1016/j.conengprac.2020.104637}
  {\path{doi:10.1016/j.conengprac.2020.104637}}.

\bibitem{RN198691}
V.~Venkatsubramanian, R.~Rengaswamy, K.~Yin, S.~N. Kavuri, A review of process
  fault detection and diagnosis part i: Quantitative model-based methods,
  Computers \& Chemical Engineering 27~(3) (2003) 293--311.
\newblock \href {https://doi.org/10.1016/s0098-1354(02)00160-6}
  {\path{doi:10.1016/s0098-1354(02)00160-6}}.

\bibitem{RN237009}
L.~Chiang, E.~Russell, R.~Braatz, Fault detection and diagnosis in industrial
  systems, Springer London, 2000.

\bibitem{RN191405}
M.~T. Amin, F.~Khan, S.~Ahmed, S.~Imtiaz, Risk-based fault detection and
  diagnosis for nonlinear and non-gaussian process systems using r-vine copula,
  Process Safety and Environmental Protection 150 (2021) 123--136.
\newblock \href {https://doi.org/10.1016/j.psep.2021.04.010}
  {\path{doi:10.1016/j.psep.2021.04.010}}.

\bibitem{RN225608}
H.~Ali, Z.~Zhang, F.~R. Gao, Multiscale monitoring of industrial chemical
  process using wavelet-entropy aided machine learning approach, Process Safety
  and Environmental Protection 180 (2023) 1053--1075.
\newblock \href {https://doi.org/10.1016/j.psep.2023.10.066}
  {\path{doi:10.1016/j.psep.2023.10.066}}.

\bibitem{RN214605}
X.~T. Bi, R.~S. Qin, D.~Y. Wu, S.~D. Zheng, J.~S. Zhao, One step forward for
  smart chemical process fault detection and diagnosis, Computers \& Chemical
  Engineering 164 (2022) 19.
\newblock \href {https://doi.org/10.1016/j.compchemeng.2022.107884}
  {\path{doi:10.1016/j.compchemeng.2022.107884}}.

\bibitem{RN198951}
L.~H. Chiang, E.~L. Russell, R.~D. Braatz, Fault diagnosis in chemical
  processes using fisher discriminant analysis, discriminant partial least
  squares, and principal component analysis, Chemometrics and Intelligent
  Laboratory Systems 50~(2) (2000) 243--252.
\newblock \href {https://doi.org/10.1016/s0169-7439(99)00061-1}
  {\path{doi:10.1016/s0169-7439(99)00061-1}}.

\bibitem{RN233663}
J.~M. Lee, C.~K. Yoo, I.~B. Lee, Statistical process monitoring with
  independent component analysis, Journal of Process Control 14~(5) (2004)
  467--485.
\newblock \href {https://doi.org/10.1016/j.jprocont.2003.09.004}
  {\path{doi:10.1016/j.jprocont.2003.09.004}}.

\bibitem{RN233068}
Z.~Q. Ge, Z.~H. Song, Process monitoring based on independent component
  analysis-principal component analysis (ica-pca) and similarity factors,
  Industrial \& Engineering Chemistry Research 46~(7) (2007) 2054--2063.
\newblock \href {https://doi.org/10.1021/ie061083g}
  {\path{doi:10.1021/ie061083g}}.

\bibitem{RN233872}
S.~J. Qin, Statistical process monitoring: basics and beyond, Journal of
  Chemometrics 17~(8-9) (2003) 480--502.
\newblock \href {https://doi.org/10.1002/cem.800} {\path{doi:10.1002/cem.800}}.

\bibitem{RN237010}
B.~Song, S.~Tan, H.~B. Shi, B.~Zhao, Fault detection and diagnosis via
  standardized k nearest neighbor for multimode process, Journal of the Taiwan
  Institute of Chemical Engineers 106 (2020) 1--8.
\newblock \href {https://doi.org/10.1016/j.jtice.2019.09.017}
  {\path{doi:10.1016/j.jtice.2019.09.017}}.

\bibitem{RN230344}
Z.~Yin, J.~Hou, Recent advances on svm based fault diagnosis and process
  monitoring in complicated industrial processes, Neurocomputing 174 (2016)
  643--650.
\newblock \href {https://doi.org/10.1016/j.neucom.2015.09.081}
  {\path{doi:10.1016/j.neucom.2015.09.081}}.

\bibitem{RN190802}
Y.~Q. Zhang, L.~Luo, X.~Ji, Y.~Y. Dai, Improved random forest algorithm based
  on decision paths for fault diagnosis of chemical process with incomplete
  data, Sensors 21~(20) (2021) 20.
\newblock \href {https://doi.org/10.3390/s21206715}
  {\path{doi:10.3390/s21206715}}.

\bibitem{RN214728}
N.~Liu, M.~G. Hu, J.~Wang, Y.~J. Ren, W.~D. Tian, Fault detection and diagnosis
  using bayesian network model combining mechanism correlation analysis and
  process data: Application to unmonitored root cause variables type faults,
  Process Safety and Environmental Protection 164 (2022) 15--29.
\newblock \href {https://doi.org/10.1016/j.psep.2022.05.073}
  {\path{doi:10.1016/j.psep.2022.05.073}}.

\bibitem{RN191413}
M.~T. Amin, F.~Khan, S.~Ahmed, S.~Imtiaz, A data-driven bayesian network
  learning method for process fault diagnosis, Process Safety and Environmental
  Protection 150 (2021) 110--122.
\newblock \href {https://doi.org/10.1016/j.psep.2021.04.004}
  {\path{doi:10.1016/j.psep.2021.04.004}}.

\bibitem{RN237011}
C.~Y. Lou, M.~A. Atoui, X.~S. Li, Novel online discriminant analysis based
  schemes to deal with observations from known and new classes: Application to
  industrial systems, Engineering Applications of Artificial Intelligence 111
  (2022) 10.
\newblock \href {https://doi.org/10.1016/j.engappai.2022.104811}
  {\path{doi:10.1016/j.engappai.2022.104811}}.

\bibitem{RN215033}
Z.~Y. Deng, T.~Han, Z.~H. Cheng, J.~J. Jiang, F.~J. Duan, Fault detection of
  petrochemical process based on space-time compressed matrix and naive bayes,
  Process Safety and Environmental Protection 160 (2022) 327--340.
\newblock \href {https://doi.org/10.1016/j.psep.2022.01.048}
  {\path{doi:10.1016/j.psep.2022.01.048}}.

\bibitem{RN237012}
Q.~C. Tao, B.~R. Xin, Y.~F. Zhang, H.~P. Jin, Q.~Li, Z.~D. Dai, Y.~Y. Dai, A
  novel triage-based fault diagnosis method for chemical process, Process
  Safety and Environmental Protection 183 (2024) 1102--1116.
\newblock \href {https://doi.org/10.1016/j.psep.2024.01.072}
  {\path{doi:10.1016/j.psep.2024.01.072}}.

\bibitem{RN194629}
H.~Wu, J.~S. Zhao, Deep convolutional neural network model based chemical
  process fault diagnosis, Computers \& Chemical Engineering 115 (2018)
  185--197.
\newblock \href {https://doi.org/10.1016/j.compchemeng.2018.04.009}
  {\path{doi:10.1016/j.compchemeng.2018.04.009}}.

\bibitem{RN195057}
Z.~P. Zhang, J.~S. Zhao, A deep belief network based fault diagnosis model for
  complex chemical processes, Computers \& Chemical Engineering 107 (2017)
  395--407.
\newblock \href {https://doi.org/10.1016/j.compchemeng.2017.02.041}
  {\path{doi:10.1016/j.compchemeng.2017.02.041}}.

\bibitem{RN237013}
S.~D. Zheng, J.~S. Zhao, A new unsupervised data mining method based on the
  stacked autoencoder for chemical process fault diagnosis, Computers \&
  Chemical Engineering 135 (2020) 17.
\newblock \href {https://doi.org/10.1016/j.compchemeng.2020.106755}
  {\path{doi:10.1016/j.compchemeng.2020.106755}}.

\bibitem{RN237014}
H.~T. Zhao, S.~Y. Sun, B.~Jin, Sequential fault diagnosis based on lstm neural
  network, IEEE Access 6 (2018) 12929--12939.
\newblock \href {https://doi.org/10.1109/access.2018.2794765}
  {\path{doi:10.1109/access.2018.2794765}}.

\bibitem{RN237015}
K.~Zhou, Y.~F. Tong, X.~T. Li, X.~R. Wei, H.~Huang, K.~Song, X.~Chen, Exploring
  global attention mechanism on fault detection and diagnosis for complex
  engineering processes, Process Safety and Environmental Protection 170 (2023)
  660--669.
\newblock \href {https://doi.org/10.1016/j.psep.2022.12.055}
  {\path{doi:10.1016/j.psep.2022.12.055}}.

\bibitem{lou2023unknown}
C.~Lou, M.~A. Atoui, Unknown health states recognition with
  collective-decision-based deep learning networks in predictive maintenance
  applications, Mathematics 12~(1) (2024) 16.
\newblock \href {https://doi.org/10.3390/math12010089}
  {\path{doi:10.3390/math12010089}}.

\bibitem{RN237035}
H.~Hashim, P.~Ryan, E.~Clifford, A statistically based fault detection and
  diagnosis approach for non-residential building water distribution systems,
  Advanced Engineering Informatics 46 (2020) 15.
\newblock \href {https://doi.org/10.1016/j.aei.2020.101187}
  {\path{doi:10.1016/j.aei.2020.101187}}.

\bibitem{RN237022}
T.~Huang, Q.~Zhang, X.~A. Tang, S.~Y. Zhao, X.~N. Lu, A novel fault diagnosis
  method based on cnn and lstm and its application in fault diagnosis for
  complex systems, Artificial Intelligence Review 55~(2) (2022) 1289--1315.
\newblock \href {https://doi.org/10.1007/s10462-021-09993-z}
  {\path{doi:10.1007/s10462-021-09993-z}}.

\bibitem{RN187493}
J.~X. Zhang, Z.~Miao, Z.~M. Feng, R.~F. Lv, C.~Y. Lu, Y.~Y. Dai, L.~C. Dong,
  Gated recurrent unit-enhanced deep convolutional neural network for real-time
  industrial process fault diagnosis, Process Safety and Environmental
  Protection 175 (2023) 129--149.
\newblock \href {https://doi.org/10.1016/j.psep.2023.05.025}
  {\path{doi:10.1016/j.psep.2023.05.025}}.

\bibitem{RN189089}
Z.~C. Wei, X.~Ji, L.~Zhou, Y.~G. Dang, Y.~Y. Dai, A novel deep learning model
  based on target transformer for fault diagnosis of chemical process, Process
  Safety and Environmental Protection 167 (2022) 480--492.
\newblock \href {https://doi.org/10.1016/j.psep.2022.09.039}
  {\path{doi:10.1016/j.psep.2022.09.039}}.

\bibitem{RN255604}
Z.~J. Ren, D.~W. Gao, Y.~S. Zhu, Q.~Ni, K.~Yan, J.~Hong, Generative adversarial
  networks driven by multi-domain information for improving the quality of
  generated samples in fault diagnosis, Engineering Applications of Artificial
  Intelligence 124 (2023) 19.
\newblock \href {https://doi.org/10.1016/j.engappai.2023.106542}
  {\path{doi:10.1016/j.engappai.2023.106542}}.

\bibitem{RN255606}
D.~W. Gao, Y.~S. Zhu, K.~Yan, H.~Fu, Z.~J. Ren, W.~Kang, C.~G. Soares, Joint
  learning system based on semi-pseudo-label reliability assessment for
  weak-fault diagnosis with few labels, Mechanical Systems and Signal
  Processing 189 (2023) 23.
\newblock \href {https://doi.org/10.1016/j.ymssp.2022.110089}
  {\path{doi:10.1016/j.ymssp.2022.110089}}.

\bibitem{lou2023recent}
C.~Lou, M.~A. Atoui, X.~Li, Recent deep learning models for diagnosis and
  health monitoring: A review of research works and future challenges,
  Transactions of the Institute of Measurement and Control (2023) 38\href
  {https://doi.org/10.1177/01423312231157118}
  {\path{doi:10.1177/01423312231157118}}.

\bibitem{RN228668}
M.~Quinones-Grueiro, A.~Prieto-Moreno, C.~Verde, O.~Llanes-Santiago,
  Data-driven monitoring of multimode continuous processes: A review,
  Chemometrics and Intelligent Laboratory Systems 189 (2019) 56--71.
\newblock \href {https://doi.org/10.1016/j.chemolab.2019.03.012}
  {\path{doi:10.1016/j.chemolab.2019.03.012}}.

\bibitem{RN237023}
S.~J. Pan, Q.~A. Yang, A survey on transfer learning, IEEE Transactions on
  Knowledge and Data Engineering 22~(10) (2010) 1345--1359.
\newblock \href {https://doi.org/10.1109/tkde.2009.191}
  {\path{doi:10.1109/tkde.2009.191}}.

\bibitem{RN216669}
H.~Wu, J.~S. Zhao, Fault detection and diagnosis based on transfer learning for
  multimode chemical processes, Computers \& Chemical Engineering 135 (2020)
  13.
\newblock \href {https://doi.org/10.1016/j.compchemeng.2020.106731}
  {\path{doi:10.1016/j.compchemeng.2020.106731}}.

\bibitem{RN192444}
Y.~L. Wang, D.~Z. Wu, X.~F. Yuan, Lda-based deep transfer learning for fault
  diagnosis in industrial chemical processes, Computers \& Chemical Engineering
  140 (2020) 13.
\newblock \href {https://doi.org/10.1016/j.compchemeng.2020.106964}
  {\path{doi:10.1016/j.compchemeng.2020.106964}}.

\bibitem{RN189748}
D.~L. Gao, X.~Z. Zhu, C.~J. Yang, X.~K. Huang, W.~H. Wang, Deep weighted joint
  distribution adaption network for fault diagnosis of blast furnace ironmaking
  process, Computers \& Chemical Engineering 162 (2022) 9.
\newblock \href {https://doi.org/10.1016/j.compchemeng.2022.107797}
  {\path{doi:10.1016/j.compchemeng.2022.107797}}.

\bibitem{RN192669}
Z.~Chai, C.~H. Zhao, A fine-grained adversarial network method for cross-domain
  industrial fault diagnosis, IEEE Transactions on Automation Science and
  Engineering 17~(3) (2020) 1432--1442.
\newblock \href {https://doi.org/10.1109/tase.2019.2957232}
  {\path{doi:10.1109/tase.2019.2957232}}.

\bibitem{RN255605}
D.~W. Gao, K.~Huang, Y.~S. Zhu, L.~B. Zhu, K.~Yan, Z.~J. Ren, C.~G. Soares,
  Semi-supervised small sample fault diagnosis under a wide range of speed
  variation conditions based on uncertainty analysis, Reliability Engineering
  \& System Safety 242 (2024) 16.
\newblock \href {https://doi.org/10.1016/j.ress.2023.109746}
  {\path{doi:10.1016/j.ress.2023.109746}}.

\bibitem{RN237017}
J.~D. Wang, C.~L. Lan, C.~Liu, Y.~D. Ouyang, T.~Qin, W.~Lu, Y.~Q. Chen, W.~J.
  Zeng, P.~S. Yu, Generalizing to unseen domains: a survey on domain
  generalization, IEEE Transactions on Knowledge and Data Engineering 35~(8)
  (2023) 8052--8072.
\newblock \href {https://doi.org/10.1109/tkde.2022.3178128}
  {\path{doi:10.1109/tkde.2022.3178128}}.

\bibitem{RN188446}
Y.~T. Xiao, H.~B. Shi, B.~Y. Wang, Y.~Tao, S.~Tan, B.~Song, Fault diagnosis of
  unseen modes in chemical processes based on labeling and class progressive
  adversarial learning, IEEE Transactions on Instrumentation and Measurement 72
  (2023) 12.
\newblock \href {https://doi.org/10.1109/tim.2022.3228271}
  {\path{doi:10.1109/tim.2022.3228271}}.

\bibitem{RN213793}
H.~Huang, R.~Wang, K.~Zhou, L.~Ning, K.~Song, Causalvit: Domain generalization
  for chemical engineering process fault detection and diagnosis, Process
  Safety and Environmental Protection 176 (2023) 155--165.
\newblock \href {https://doi.org/10.1016/j.psep.2023.06.018}
  {\path{doi:10.1016/j.psep.2023.06.018}}.

\bibitem{RN190479}
Y.~T. Xiao, H.~B. Shi, B.~Y. Wang, Y.~Tao, S.~Tan, B.~Song, Weighted
  conditional discriminant analysis for unseen operating modes fault diagnosis
  in chemical processes, Ieee Transactions on Instrumentation and Measurement
  71 (2022) 14.
\newblock \href {https://doi.org/10.1109/tim.2022.3152235}
  {\path{doi:10.1109/tim.2022.3152235}}.

\bibitem{RN187888}
C.~Zhao, W.~M. Shen, Adversarial mutual information-guided single domain
  generalization network for intelligent fault diagnosis, IEEE Transactions on
  Industrial Informatics 19~(3) (2023) 2909--2918.
\newblock \href {https://doi.org/10.1109/tii.2022.3175018}
  {\path{doi:10.1109/tii.2022.3175018}}.

\bibitem{RN186301}
J.~Wang, H.~Ren, C.~Q. Shen, W.~G. Huang, Z.~K. Zhu, Multi-scale style
  generative and adversarial contrastive networks for single domain
  generalization fault diagnosis, Reliability Engineering \& System Safety 243
  (2024) 13.
\newblock \href {https://doi.org/10.1016/j.ress.2023.109879}
  {\path{doi:10.1016/j.ress.2023.109879}}.

\bibitem{RN237026}
Y.~Y. Pu, J.~Tang, X.~G. Li, C.~Wei, W.~B. Huang, X.~X. Ding, Single-domain
  incremental generation network for machinery intelligent fault diagnosis
  under unknown working speeds, Advanced Engineering Informatics 60 (2024) 14.
\newblock \href {https://doi.org/10.1016/j.aei.2024.102400}
  {\path{doi:10.1016/j.aei.2024.102400}}.

\bibitem{RN237028}
Y.~Guo, J.~Zhang, Chemical fault diagnosis network based on single domain
  generalization, Process Safety and Environmental Protection 188 (2024)
  1133--1144.
\newblock \href {https://doi.org/https://doi.org/10.1016/j.psep.2024.05.106}
  {\path{doi:https://doi.org/10.1016/j.psep.2024.05.106}}.

\bibitem{RN237024}
L.~A. Gatys, A.~S. Ecker, M.~Bethge, Ieee, Image style transfer using
  convolutional neural networks, in: 2016 IEEE Conference on Computer Vision
  and Pattern Recognition (CVPR), IEEE Conference on Computer Vision and
  Pattern Recognition, Ieee, NEW YORK, 2016, pp. 2414--2423.
\newblock \href {https://doi.org/10.1109/cvpr.2016.265}
  {\path{doi:10.1109/cvpr.2016.265}}.

\bibitem{RN237029}
X.~Glorot, A.~Bordes, Y.~Bengio, Deep sparse rectifier neural networks, in:
  Proceedings of the Fourteenth International Conference on Artificial
  Intelligence and Statistics, JMLR Workshop and Conference Proceedings, 2011,
  pp. 315--323.

\bibitem{RN255602}
Y.~Zhang, W.~Li, W.~Sun, R.~Tao, Q.~Du, Single-source domain expansion network
  for cross-scene hyperspectral image classification, IEEE Transactions on
  Image Processing 32 (2023) 1498--1512.
\newblock \href {https://doi.org/10.1109/TIP.2023.3243853}
  {\path{doi:10.1109/TIP.2023.3243853}}.

\bibitem{RN237019}
M.~Long, Z.~Cao, J.~Wang, M.~I. Jordan, Conditional adversarial domain
  adaptation, Advances in Neural Information Processing Systems 31 (2018).

\bibitem{RN237018}
P.~Khosla, P.~Teterwak, C.~Wang, A.~Sarna, Y.~Tian, P.~Isola, A.~Maschinot,
  C.~Liu, D.~Krishnan, Supervised contrastive learning, Advances in Neural
  Information Processing Systems 33 (2020) 18661--18673.

\bibitem{RN237020}
J.~J. Downs, E.~F. Vogel, A plant-wide industrical-process control problem,
  Computers \& Chemical Engineering 17~(3) (1993) 245--255.
\newblock \href {https://doi.org/10.1016/0098-1354(93)80018-i}
  {\path{doi:10.1016/0098-1354(93)80018-i}}.

\bibitem{RN237031}
K.~E.~S. Pilario, Y.~Cao, Canonical variate dissimilarity analysis for process
  incipient fault detection, IEEE Transactions on Industrial Informatics
  14~(12) (2018) 5308--5315.
\newblock \href {https://doi.org/10.1109/tii.2018.2810822}
  {\path{doi:10.1109/tii.2018.2810822}}.

\bibitem{RN237021}
A.~Bathelt, N.~L. Ricker, M.~Jelali, Revision of the tennessee eastman process
  model, IFAC-PapersOnLine 48~(8) (2015) 309--314.
\newblock \href {https://doi.org/10.1016/j.ifacol.2015.08.199}
  {\path{doi:10.1016/j.ifacol.2015.08.199}}.

\bibitem{RN237030}
Z.~Y. Liu, C.~Li, X.~He, Evidential ensemble preference-guided learning
  approach for real-time multimode fault diagnosis, IEEE Transactions on
  Industrial Informatics 20~(4) (2024) 5495--5504.
\newblock \href {https://doi.org/10.1109/tii.2023.3332112}
  {\path{doi:10.1109/tii.2023.3332112}}.

\end{thebibliography}

\section*{Code Availability}
\href{https://github.com/GuangqiangLi/DACN}{https://github.com/GuangqiangLi/DACN} (it will be available after being published.)











\end{document}